\documentclass{article}

\usepackage{arxiv}

\usepackage[utf8]{inputenc} % allow utf-8 input
\usepackage[T1]{fontenc}    % use 8-bit T1 fonts
\usepackage{hyperref}       % hyperlinks
\usepackage{url}            % simple URL typesetting
\usepackage{booktabs}       % professional-quality tables
\usepackage{amsfonts}       % blackboard math symbols
\usepackage{nicefrac}       % compact symbols for 1/2, etc.
\usepackage{microtype}      % microtypography
\usepackage{lipsum}		% Can be removed after putting your text content
\usepackage{graphicx}
\usepackage{doi}

\usepackage[
    backend=biber,
    style=alphabetic,
    sorting=nyt,
    maxbibnames=99
    ]{biblatex}
\usepackage{acronym}        	% Added to support acronyms
\usepackage[inline]{enumitem}   % Inline enumerate
\usepackage{todonotes}	        % Todo-notes - Use [disable] to make them disappear
\usepackage{xfrac}              % Other fraction forms
\usepackage{amssymb}            % Cross and tick
\usepackage{pifont}             % Cross and tick
\usepackage{graphicx}           % 2x2 graphics
\usepackage{subcaption}         % 2x2 graphics
\usepackage{mwe}                % 2x2 graphics
\usepackage{amsmath}            % Argmin argument
\usepackage[toc,page]{appendix} % Proper referencing to the Appendix
\usepackage[left]{lineno}
\newcommand{\cmark}{\ding{51}}
\newcommand{\xmark}{\ding{55}}
    
\newacro{BL}[BL]{Baseline}
\newacro{ETA}[ETA]{Estimated Time of Arrival}
\newacro{LIME}[LIME]{Local Interpretable Model-agnostic Explanations}
\newacro{MAE}[MAE]{Mean Absolute Error}
\newacro{MRE}[MRE]{Mean Relative Error}
\newacro{MAPE}[MAPE]{Mean Absolute Percentage Error}
\newacro{ML}[ML]{Machine Learning}
\newacro{MLR}[MLR]{Multiple Linear Regression}
\newacro{NN}[FCNN]{Fully-Connected Feedforward Neural Network}
\newacro{RF}[RF]{Random Forest}
\newacro{SHAP}[SHAP]{Shapley Additive Explanations}
\newacro{XAI}[XAI]{eXplainable Artificial Intelligence}

\DeclareMathOperator*{\argmin}{arg\,min}

\title{An Explainable Stacked Ensemble Model for Static Route-Free Estimation of Time of Arrival}

\author{ \href{https://orcid.org/0000-0001-7181-5336}{\includegraphics[scale=0.06]{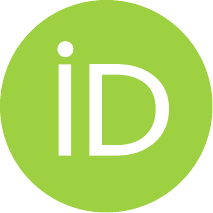}\hspace{1mm}Sören Schleibaum}\\
	Department of Informatics\\
	Clausthal University of Technology\\
	Julius-Albert-Straße 4 in 38678 Clausthal-Zellerfeld, Germany\\
	\texttt{soeren.schleibaum@tu-clausthal.de} \\
	\And
	\href{https://orcid.org/0000-0001-7533-3852}{\includegraphics[scale=0.06]{orcid.pdf}\hspace{1mm}Jörg P. Müller}\\
	Department of Informatics\\
	Clausthal University of Technology\\
	Julius-Albert-Straße 4 in 38678 Clausthal-Zellerfeld, Germany\\
	\texttt{joerg.mueller@tu-clausthal.de} \\
	\And
	\href{https://orcid.org/0000-0002-6656-8809}{\includegraphics[scale=0.06]{orcid.pdf}\hspace{1mm}Monika Sester}\\
	Institute of Cartography and Geoinformatics\\
	Leibniz University Hanover\\
	Appelstraße 9a in 30167 Hanover, Germany\\
	\texttt{monika.sester@ikg.uni-hannover.de}
}

\renewcommand{\shorttitle}{An Explainable Stacked Ensemble Model for Static Route-Free \acs{ETA}}

\hypersetup{
    pdftitle={An Explainable Stacked Ensemble Model for Static Route-Free Estimation of Time of Arrival},
    pdfauthor={S. Schleibaum, J. P. Müller, and M. Sester},
    pdfkeywords={Estimated Time of Arrival, Ensemble Learning, eXplainable Artificial Intelligence},
}

\begin{document}
\maketitle

\begin{abstract}
Sustainable concepts for on-demand transportation, such as ride-sharing or ride-hailing, require advanced technologies and novel dynamic planning and prediction methods. In this paper, we consider the prediction of taxi trip durations, focusing on the problem of \ac{ETA}. \ac{ETA} can be used to compute and compare alternative taxi schedules and to provide information to drivers and passengers. To solve the underlying hard computational problem with high precision, \ac{ML} models for \ac{ETA} are state of the art. However, these models are mostly \textit{black-box} neural networks. Hence, the resulting predictions are difficult to explain to users. To address this problem, the contributions of this paper are threefold. First, we propose a novel stacked \textit{two-level ensemble model} combining multiple \ac{ETA} models; we show that the stacked model outperforms state-of-the-art \ac{ML} models. However, the complex ensemble architecture makes the resulting predictions less transparent. To alleviate this, we investigate \ac{XAI} methods for explaining the first- and second-level models of the ensemble. Third, we consider and compare different ways of combining first-level and second-level explanations. This novel concept enables us to explain stacked ensembles for regression tasks. The experimental evaluation indicates that the considered \ac{ETA} models correctly learn the importance of those input features driving the prediction.

\emph{Paper is accepted at the \hyperlink{https://www.hindawi.com/journals/jat/}{Journal of Advanced Transportation from Hindawi}.}
\end{abstract}


\section{Introduction}
\label{sec:Introduction}
\begin{figure}[ht]
    \centering
    \includegraphics[width=0.75\textwidth]{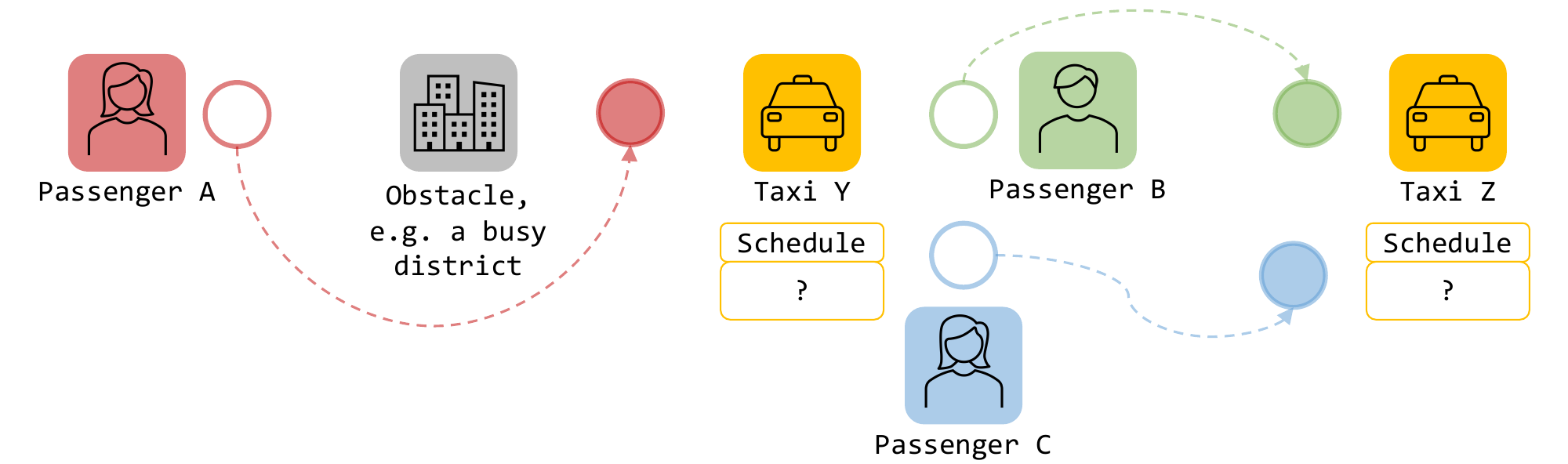}
    \caption{Motivating scenario about \ac{ETA} for the planning of taxi schedules.}
    \label{fig:sampleETAMotivation}
\end{figure}

In intelligent transportation systems for fleet coordination and optimization (e.g., a ridesharing service), the computation and optimization of taxi schedules are often supported by a component that estimates the duration or time of arrival for a given trip. To illustrate the problem of \ac{ETA}, in Figure~\ref{fig:sampleETAMotivation}, we show two taxis $Y$ and $Z$ that aim to serve three passengers $A$, $B$, and $C$. Even in this small example, different alternative schedules are to be considered in order to find a close-to-optimal one. Using an algorithm for \ac{ETA} that is independent of a route, it is possible to avoid having to compute all routes in advance, which leads to considerable speed-up for larger and dynamic problem instances. 

\ac{ETA} also helps provide models for predicting upcoming taxi trips, e.g. \textit{when} a taxi will pick up a passenger or \textit{how long} a trip will take for a driver/passenger. State-of-the-art approaches have shown that high prediction precision can be achieved using \ac{ML} \cite{dearaujo.2019, kankanamge.2019, li.2018, schleibaum.2022}. A promising option to further increase prediction precision are ensemble models \cite{ganaie.2022}; a special type of ensemble models is a stacked ensemble: here, the output of multiple first-level models is combined via another (second-level) model \cite{ganaie.2022}, for example to the final estimation of the \ac{ETA} for a taxi trip. The higher variety achieved via multiple models, potentially of different types, can better represent and interpret the diversity of the data and potentially increase prediction precision. One drawback is that by combining several black-box models with a second-level black box, the resulting model becomes even less transparent. This means that it is very difficult to understand why the model proposes a certain solution. One option to remedy this drawback is to apply \ac{XAI}-methods \cite{linardatos.2020,lundberg.2017} like \ac{SHAP}, which aim to explain the output of complex non-linear \ac{ML} models like neural networks. With such \ac{XAI} methods, we are able to learn the influence of input features on an estimated trip duration, e.g. ``\emph{The hour (8:00 am) increases the estimated trip duration by 25 seconds.}''

The main contributions of this paper are:
\begin{enumerate}
    \item Inspired by the good result for stacked ensembles in other problems \cite{xia.2021,gour.2022,akhtar.2020}, we propose and evaluate a stacked ensemble model for route-free estimation of trip durations. 
    \item We enable explainability of stacked ensemble structures by extending existing \ac{XAI} methods based on feature importance; in particular, we propose and compare three novel joining methods.
\end{enumerate}

The paper is organized as follows. In Section~\ref{sec:RelatedWork}, we derive our research gap and aim based on a related work review on route-free \ac{ETA} and the explanation of stacked ensembles. In Section~\ref{sec:Methodology}, we describe preliminaries, the data sets used to evaluate the prediction precision of the stacked ensemble, the first level \ac{ETA} models, the selection of the \ac{XAI} methods, and the evaluation procedure. Subsequently, the construction of the stacked ensemble model is described in Section~\ref{sec:ModelsForETA}. In Section~\ref{sec:ExplainingFirstLevelModelsForETA}, we select and apply state-of-the-art \ac{XAI} methods to explain the first-level models. Then, we propose the joining methods to explain the ensemble and conduct simulation experiments to evaluate our approach in Section~\ref{sec:ExplainingEnsembleModelsForETA}. Section~\ref{sec:Discussion} discusses our findings with respect to the research aim and points out limitations as well as venues for future work; we conclude the paper in Section~\ref{sec:Conclusion}. The source code used in this paper can be accessed online via \cite{codeeta.2022}.

\section{Related Work}
\label{sec:RelatedWork}

\subsection{Route-Free Estimated Time of Arrival}
We list all the considered works about route-free \ac{ETA} in Table~\ref{tab:RelatedWorkETA}. \cite{jindal.2018} and \cite{haliem.2021a}, develop a route-free \ac{ETA} approach as part of a larger ridesharing service. First, \textcite{jindal.2018} estimate a trip's distance and then its \ac{ETA}--both via \ac{NN}s. \textcite{haliem.2021a} tackle \ac{ETA} based on a single \ac{NN} with two hidden layers. While both use a relatively simple network architecture, they achieve remarkable prediction precision.

Among the approaches that focus on \ac{ETA}, \textcite{wang.2019} propose a relatively simple neighbor-based method. Similar to \cite{jindal.2018,haliem.2021a}, \cite{dearaujo.2019} use a \ac{NN}, but with a different architecture. After searching for the best representation of the pickup and dropoff location that is passed into an \ac{ETA} model, \textcite{schleibaum.2022} propose another \ac{NN} architecture. \textcite{tagelsir.2022} develop an advanced deep learning-based system; they incorporate both, spatial-temporal and external features, via convolutional, fully connected, and attention layers. 

Similar to \cite{jindal.2018}, which use an ensemble of two networks for two different tasks, \textcite{zou.2020} propose a stacked ensemble of a gradient boosting decision tree and a fully-connected feed-forward neural network to estimate the time of arrival; both first level models consume the same feature set. 

\begin{table}[ht]
	\caption{Related work on route-free \ac{ETA}}
	\begin{tabular*}{\hsize}{@{\extracolsep{\fill}}lcccc@{}}
		\toprule
		Reference              & Usage of \ac{ML} model(s) & Usage of ensemble & Ensemble type & Explanation \\
		\midrule
		\cite{jindal.2018}     & \cmark               & (\cmark)          & Stacked       & \xmark  \\ % FCNN
        \cite{haliem.2021a}    & \cmark               & \xmark            & -             & \xmark  \\ % FCNN
        \cite{wang.2019}       & \xmark               & \xmark            & -             & \xmark  \\ % Neighbor-based
        \cite{dearaujo.2019}   & \cmark               & \xmark            & -             & \xmark  \\ % FCNN
        \cite{schleibaum.2022} & \cmark               & \xmark            & -             & \xmark  \\ % FCNN
        \cite{tagelsir.2022}   & \cmark               & \xmark            & -             & \xmark  \\ % Neural network
        \cite{zou.2020}        & \cmark               & \cmark            & Same-level    & \xmark  \\ % Neural network
		\bottomrule
	\end{tabular*}
	\label{tab:RelatedWorkETA}
\end{table}

\subsection{Explaining Ensembles}
As shown in Table~\ref{tab:RelatedWorkXAIAndEnsembles}, except for \cite{deng.2019,juraev.2022,ahmed.2021}, all works focusing on explaining ensembles only tackle a classification problem. The majority of these works--\cite{bologna.2018,bologna.2021,obregon.2023,sendi.2019}--explain the ensembles post-hoc by extracting rules. Given an ensemble of homogeneous models and homogeneous feature sets, \textcite{bologna.2018} propose to transform the models of the ensemble into discretized interpretable multilayer perceptrons--a neural network derivative. From this new ensemble, rules are extracted as an explanation. In a more recent work, \textcite{bologna.2021} propose another method to extract rules from same-level ensembles. \textcite{sendi.2019} learn a same-level ensemble of neural networks, transform it into one of decision trees, and use a multi-agent dialog approach to extract relatively simple rules to explain the learned classification pattern. Recently \textcite{obregon.2023} also proposed to extract simple rules from an ensemble by combining and simplifying their base trees. 

Those works that do not use rule extraction to explain ensembles for classification are \cite{silva.2019,kallipolitis.2021}. \textcite{khalifa.2022} propose a method to transform a learned ensemble of decision trees into a single decision tree; although they limit the prediction precision of their ensemble by ceiling the depth of their decision trees, their simplified tree remains the same prediction precision as the ensemble is remarkable. To explain a stacked ensemble for classification, \textcite{silva.2019} present the results of several \ac{XAI} methods--text-based rules extracted from a decision tree, feature importance from scorecards, and an example-based method--beside each other; the authors apply their explanation approach to several ensembles used in medicine and finance. 

Similarly to \cite{silva.2019}, \textcite{ren.2022} predict the survival rate of patients and apply the \ac{XAI} method SHAP to determine the contributions of the input features for a stacked ensemble. Also in the field of medicine, \textcite{ahmed.2021} apply several \ac{XAI} methods to an ensemble that predicts the mortality rate of patients. 

Both--\cite{deng.2019} and \cite{juraev.2022}--explain an ensemble independent from the tackled problem. While \textcite{deng.2019} also extract rules from a same-level ensemble of decision trees, \textcite{juraev.2022} apply SHAP to a stacked ensemble that does both, classification and regression.

\begin{table}[ht]
    \small
	\caption{Related work that explains ensembles}
	\begin{tabular*}{\hsize}{@{\extracolsep{\fill}}llllll@{}}
		\toprule
		Reference                 & Tackled problem               & Ensemble type     & Model types       & Feature sets      & Explanation type      \\
		\midrule
		\cite{ahmed.2021}        & Regression                     & Stacked           & Heterogeneous     & Homogeneous       & Global, post-hoc           \\
        \cite{bologna.2018}      & Classification                 & Same-level        & Homogeneous       & Homogeneous       & Global, post-hoc           \\
        \cite{bologna.2021}      & Classification                 & Same-level        & Homogeneous       & Homogeneous       & Global, post-hoc           \\
        \cite{deng.2019}         & Classification and regression  & Same-level        & Homogeneous       & Homogeneous       & Global, post-hoc           \\
        \cite{juraev.2022}       & Classification and regression  & Stacked           & Heterogeneous     & Heterogeneous     & Local/global, post-hoc \\
        \cite{kallipolitis.2021} & Classification                 & Same-level        & Homogeneous       & Homogeneous       & Local, post-hoc            \\
        \cite{khalifa.2022}      & Classification                 & Same-level        & Homogeneous       & Homogeneous       & Global, post-hoc           \\
        \cite{obregon.2023}      & Classification                 & Same-level        & Homogeneous       & Homogeneous       & Local, post-hoc            \\
        \cite{ren.2022}          & Classification                 & Stacked           & Heterogeneous     & Homogeneous       & Global, post-hoc           \\
        \cite{sendi.2019}        & Classification                 & Same-level        & Homogeneous       & Homogeneous       & Local, post-hoc            \\
        \cite{silva.2019}        & Classification                 & Stacked           & Heterogeneous     & Homogeneous       & Local, post-hoc            \\
		\bottomrule
	\end{tabular*}
	\label{tab:RelatedWorkXAIAndEnsembles}
\end{table}

\subsection{Research Gap and Aim}
In general, we observe that the comparability of the aforementioned route-free \ac{ETA} approaches is complicated due to the varying evaluation metrics applied, the different data sets used, and the diverse feature sets selected. Even though the feature sets selected seem to depend on the architecture applied and the promising results of \cite{zou.2020}, no previous approach tried a stacked ensemble with heterogeneous feature sets at the first level. 

Except for \cite{wang.2019}, all of the aforementioned approaches proposed \ac{ML}-based models for \ac{ETA}. Even though such complex models are known to learn intricate patterns in the input data, none made the learned patterns transparent through explanations. Furthermore, explaining a stacked ensemble for regression is not straightforward. While most related work focuses on same-level ensembles, only four works explain a stacked ensemble--\cite{ahmed.2021,juraev.2022,ren.2022,silva.2019}--and all through existing \ac{XAI} methods. While \cite{ahmed.2021,ren.2022} approach the ensemble as one model, \cite{juraev.2022,silva.2019} explain each first-level model of the ensemble separately. Both approaches hide the contribution of single models to the final decision. 

Consequently, we conclude that the combination of multiple models to a \emph{stacked ensemble to perform \ac{ETA}} and \emph{explaining both--the classical single-level models and the stacked ensemble}--is an open research gap. Our research aim is twofold: First, we form a stacked ensemble model to tackle \ac{ETA}. Second, we aim for an explanation method to explain such a stacked ensemble. As the explanation of ensembles through the extraction of rules is common, we will focus on feature importance methods. To limit the scope of this paper, we focus on local post-hoc explanations.

\section{Materials and Methods}
\label{sec:Methodology}

\subsection{Preliminaries}
\label{sec:Preliminaries}

\paragraph{Estimating the Time of Arrival.} Given a potential trip represented by a set of features $X=\{x_1, x_2, \ldots, x_n\}$ such as the latitude and longitude of its starting location, \ac{ETA} aims to predict its duration $\hat{y} \in \mathbb{R}$ by a function $f$ so that $f(X)=\hat{y}$. The goal is to find an $f$ that minimizes the difference between $\hat{y}$ and the real duration $y$. As $y$ is continuous, the problem described is a classical regression problem. As shown above, most of the related work uses deep learning to learn $f$ based on a set of historical trips and their durations. Because we consider route-free \ac{ETA}, information about the route, such as the number of turns on the route, or information not known before a trip starts, such as a traffic accident on the route happening after the start of a trip, are excluded from $X$.

\begin{figure}[ht]
    \centerline{\includegraphics[width=.4\textwidth]{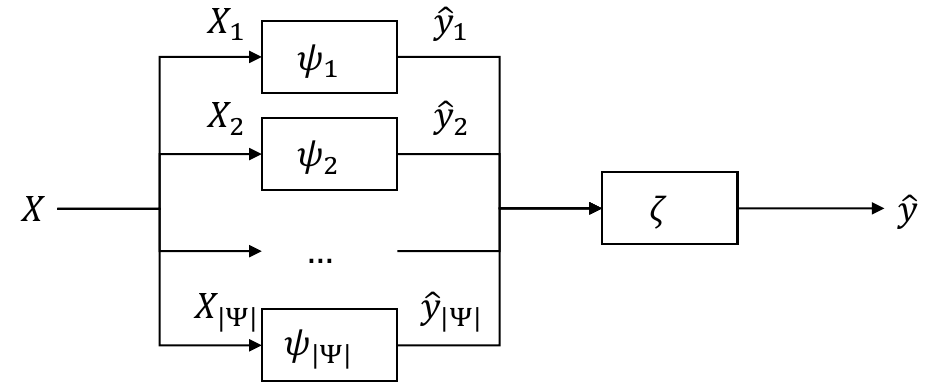}}
    \caption{Architecture of a stacked ensemble with two levels--the models $\psi_1, \psi_2, \ldots, \psi_{|\Psi|}$ build the first level and the model $\zeta$ the second level.}
    \label{fig:EnsembleIllustration}
\end{figure}

\paragraph{Ensemble Learning.} The function $f$ that tackles the aforementioned \ac{ETA} problem can be realized through a stacked ensemble with two levels. On the first level, such an ensemble is a composition of multiple functions $\psi_i \in \Psi$. Each $\psi_i$ estimates $f$ based on its feature subset $X_i \subseteq X$ so that $\psi_i(X_i) = \hat{y}_i, \forall \psi_i \in \Psi$. On the second level of the ensemble, another function $\zeta$ estimates $y$ based on the outputs of the first level so that $\zeta(\hat{y}_1, \hat{y}_2, \ldots, \hat{y}_{|\Psi|}) = \hat{y}$. As we focus on a heterogeneous ensemble, we require the models that realize the functions $\psi_i \in \Psi$ to be of different types--like a tree- and a neural network-based model. An illustration of the ensemble architecture is shown in Figure~\ref{fig:EnsembleIllustration}. 

\paragraph{Explanations.} Throughout this paper, we consider an explanation as a vector $e$ of length $|X|$ assigning a value to each input feature $x \in X$ for a given prediction $\hat{y}$: $e = {e_1, e_2, \cdots, e_x}$; so, $e \in \mathbb{R}^{|X|}$. As we consider the prediction as given, we explain the model post hoc. To differentiate the explanations in an ensemble, we add a model superscript to an explanation: $e^M$.

\subsection{Data Sets}
\label{sec:Dataset}

\paragraph{Selection and Features.} We select two data sets: the New York City Yellow taxi trip data from 2015 and 2016--see \cite{newyork.2022}--and one recorded in Washington DC in 2017--see \cite{kaggle.2019}. We select the former because it was used several times to demonstrate an \ac{ETA} approach and it is the data set used mainly in this paper. We additionally include the Washington DC data set to increase the generalizability of our experiment as regards the usage of ensembles to increase the prediction precision. For both data sets, we rely on the feature engineering described by \textcite{schleibaum.2022}, which makes use of or enhances the data set by the following features: the location-based ones with
\begin{enumerate*}[label=(\arabic*)]
    \item the pickup/dropoff as degree-based coordinates and 
    \item the indices of a 50-meter square grid as an alternative representation. To represent the start time of a trip,
    \item the \emph{month}, 
    \item \emph{week}, 
    \item \emph{weekday}, and
    \item the indices of a \emph{5-minute time-bin}, which represents the hour and minute, are used. Moreover, we use
    \item the \emph{temperature} at the hour a trip starts and calculate
    \item the \emph{Haversine distance} between pickup and dropoff location.
\end{enumerate*}

\begin{figure}
    \centering
    \begin{subfigure}[b]{0.49\textwidth}
        \centering
        \includegraphics[width=\textwidth]{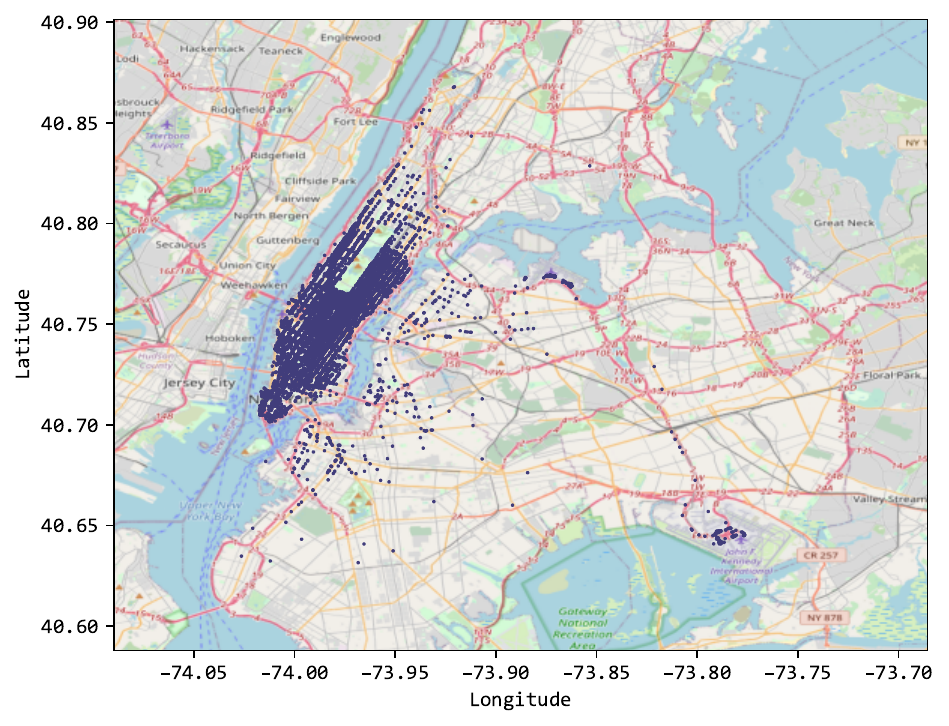}
        \caption{}
        \label{fig:pickupDistribution}
    \end{subfigure}
    \hfill
    \begin{subfigure}[b]{0.49\textwidth}
        \centering
        \includegraphics[width=\textwidth]{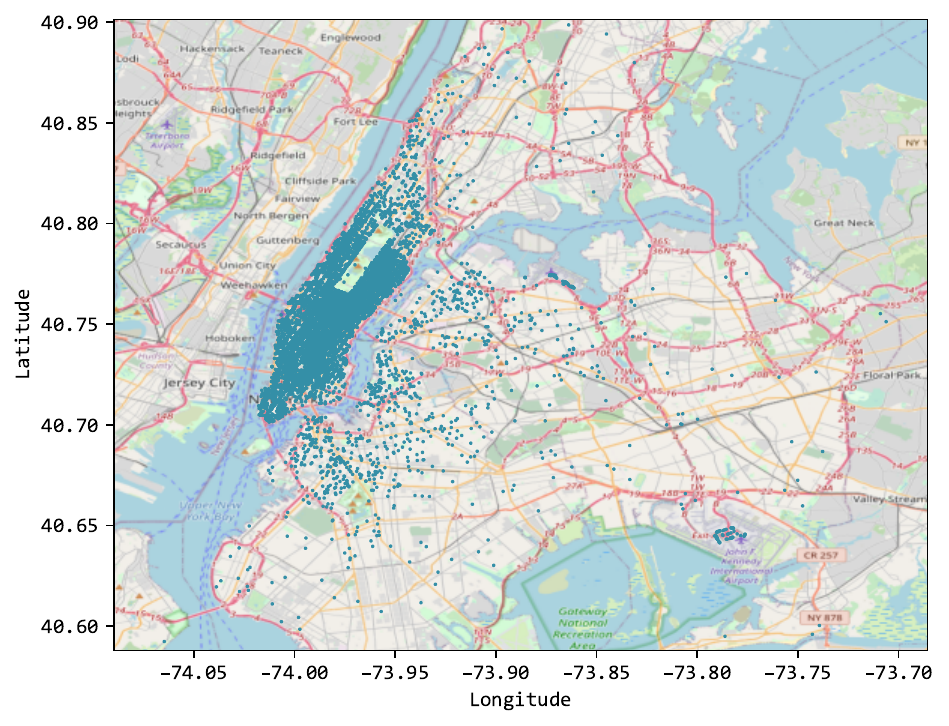}
        \caption{}
        \label{fig:dropoffDistribution}
    \end{subfigure}
    \caption{Distribution of the pickup (a) and dropoff (b) locations for randomly selected trips from the training data of the New York City data set.}
    \label{fig:pickupDropoffDistributions}
\end{figure}

\paragraph{Outlier Removal.} For removing outliers, we also use the criteria from \textcite{schleibaum.2022} and the description of the following method partly reproduces their wording. Overall, around 3\% of the trips from the New York City data set and around 19\% percent from the Washington DC data set are filtered out. A trip can be an outlier because one of its locations is not in the area studied, which is shown in Figure~\ref{fig:pickupDropoffDistributions}, or not in a district like erroneously being recorded in the Hudson River. Moreover, a trip`s reported duration could be unreasonably low or high, or could not be logically correlated with the distance between pickup and dropoff locations; we also remove trips with a distance of zero. Compared to other papers, the criteria are relatively moderate and, therefore, the comparison to approaches not reproduced is fairer.

\begin{figure}
    \centerline{\includegraphics[width=\textwidth]{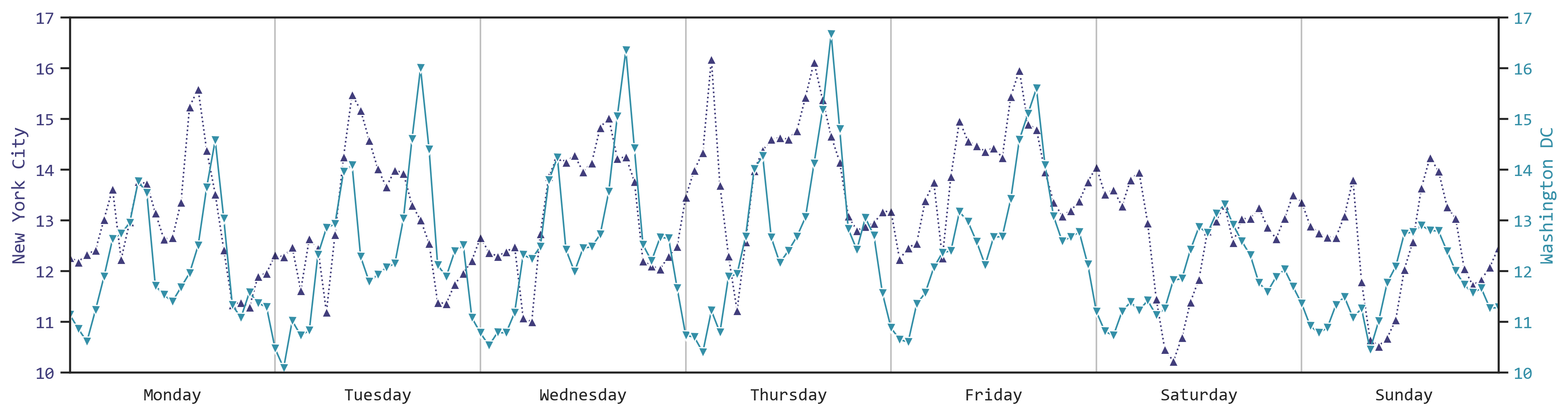}}
    \caption{Distribution of the average duration in minutes over the week in the New York City data set (darker blue) and in the Washington DC data set (lighter blue).}
    \label{fig:NofTripsAndD)urationVsTimeOfWeekday}
\end{figure}

\paragraph{Characteristics.} To better understand the data, in Figure~\ref{fig:NofTripsAndD)urationVsTimeOfWeekday}, we visualize the average duration of the trip per weekday for both data sets. In both, the number of trips is relatively low during the early morning; during the week, it has one peak at around 8~am and another one at around 5~pm. 
As expected, during the weekend, the morning peak does not exist; on average, the average duration of trips is lower than during the week.

To show the area considered and to better understand the distributions of pickup and dropoff locations, we visualize both in Figure~\ref{fig:pickupDropoffDistributions}. As expected for the Yellow taxis in New York City, the vast majority of trips start in Manhattan; most of the trips not starting in Manhattan begin at John F. Kennedy Airport. As for the dropoff locations, the general behavior is similar, but more trips end outside of Manhattan.

\subsection{Estimated Time of Arrival Models} 
We take the three \ac{ETA} models proposed by \textcite{schleibaum.2022} and their hyperparameters as our first-level models. We chose these because they are sophisticated \ac{ML} methods previously used for tackling the problem of static route-free \ac{ETA}. Further, these models are based on bagging (learning multiple models from different subsets of a data set), boosting (learning multiple models sequentially based on the previous models), and neural networks (nodes stacked in several layers enabling the capturing of complex patterns especially when trained on large data sets). Thereby, the three \ac{ML} methods \ac{RF}, XGBoost, and a neural network based on three diverse concepts provide a good basis for a heterogeneous ensemble. 

As alternatives for the second-level model, we consider the same \ac{ML} methods and add a relatively simple \ac{MLR}. Regarding the main data set or the one from New York City, we use 1M trips for training and validation from 2015 and another 250K from 2016 for testing. For the data set from Washington DC, we use less or 600K trips for training and validation, and another 50K for testing. As training data for the second-level models, we use the predictions of the first-level models on the validation data and use the same test data as before. We do not use the same training data twice or for the first- and second-level models to reduce overfitting. We do not tune the hyperparameters of the second-level models and, therefore, consider not using a validation data set as fine. Except for the \ac{MLR}, which does not have any hyperparameters, for the second-level models, we use the same hyperparameters as for the first-level models. The only difference is that we decrease the number of trees for the \ac{RF} and XGBoost from 300 to 100 and the number of hidden layers for the neural network-based model from four to two; we choose smaller models compared to the first-level models because the number of input features or the variety of the model input is reduced substantially.

\paragraph{Baselines.} We select three of the approaches presented in Table~\ref{tab:RelatedWorkETA} reproducible from the corresponding paper as baselines: \cite{jindal.2018}, \cite{dearaujo.2019}, \cite{haliem.2021a}. For all three, we perform hyperparameter tuning via a random grid search. In particular, we tune the learning rate and batch size. We perform early stopping with a patience of 30 epochs. While we use the \ac{MAE} for optimizing \cite{jindal.2018} and \cite{haliem.2021a}, the mean squared error is used for reproducing \cite{dearaujo.2019} as described in their paper.

\subsection{Selection of \acs{XAI} Methods} 
To demonstrate our approach, we select two commonly used \ac{XAI} methods--\ac{LIME} and \ac{SHAP}--which are described below. We chose these \ac{XAI} methods because both are model-agnostic and can, therefore, be applied to all models of the heterogeneous ensemble. Moreover, both create local post hoc explanations that can be used to explain to \ac{ETA} users such as taxi drivers and passengers. Although all first-level models are explained via the \ac{XAI} methods, only the second-level model that performs the best will be explained.

\paragraph{\acl{LIME}.} \textcite{ribeiro.2016} present \ac{LIME}, which explains predictions based on a linear surrogate model by minimizing two aspects: the goodness of the local approximation of the interpreted model in the observations neighborhood and the complexity of the surrogate model. This post-hoc \ac{XAI} method outputs a vector- or graphics-based explanation that is visualized differently by software libraries. The main formula presented by \textcite{ribeiro.2016} is:
\begin{equation} 
    \xi(x) = \argmin_{g \in G} \mathcal{L}(f, g, \pi_x) + \Omega(g)
\end{equation}
The importance of feature $x$ from a sample is extracted from the surrogate model $g$ from all possible surrogate models $G$ that describe the black box model $f$ and the neighborhood of $x$--denoted as $\pi_x$--best, while also minimizing the complexity of the surrogate model $\Omega(g)$.

\paragraph{\acl{SHAP}.} Another model-agnostic \ac{XAI}-method--\ac{SHAP}--was proposed by \textcite{lundberg.2017}. It is able to generate local explanations for a given sample by making the features' importance transparent. Therefore, SHAP utilizes the famous Shapley values from cooperative game theory. More concretely, \textcite{lundberg.2017} presented the following formula:    
\begin{equation} 
\phi_i = \sum_{S \subseteq F \setminus \{i\}}  \frac{\vert S \vert ! (\vert F \vert - \vert S \vert - 1)!}{\vert F \vert !} 
    [f_{S \cup \{i\}} (x_{S \cup \{i\}}) - f_S(x_S)]
\end{equation}
Here, the contribution of a feature $\phi_i$ is estimated by iterating over all subsets of features $S$ of the feature set $F$ without the feature $i$. The fraction in the sum weights the difference between the output of the model to be explained--represented by the function $f$--with and without $i$ or the contribution of $i$.

\subsection{Evaluation}

\paragraph{Prediction Precision of \ac{ETA} models.} Similar to \textcite{schleibaum.2022}, we apply three evaluation metrics common for regression tasks: 
\begin{enumerate*}[label=(\arabic*)]
    \item The \acl{MAE} (\({MAE = \sfrac{1}{N} \sum_{i} \vert y_i - \hat{y}_i \vert }\)), which in our case returns the average error per trip in seconds, 
    \item the \acl{MRE} (\({ MRE = \sum_{i} |y_i - \hat{y}_i| / \sum_{i} y_i }\)), and 
    \item the \acl{MAPE} (\({ MAPE = \sfrac{1}{N} \sum_{i} \left| (y_i - \hat{y}_i)/y_i \right| }\)), which is robust to outliers.
\end{enumerate*}
Because the latter two produce percentage values, they are also relatively easy to understand and put the error in perspective to a trip's duration.

\paragraph{Scenarios for Explanations.} To demonstrate and evaluate our explanation approach, we randomly select ten trips from the New York City test data for four scenarios. Each scenario has two opposing characteristics that are described in the following together with the scenarios: 
\begin{enumerate}[label=SC\arabic* - ]
    \item \emph{Off city-center vs. city-center}: We compare trips that start outside of the city-center--a rectangle with the bottom left at coordinate (40.7975, -73.9619) and top right at (40.8186, -73.9356)--with those that do start in the city-center--a rectangle with the bottom left at (40.7361, -73.9980) and top right at (40.7644, -73.9770).
    \item \emph{Night-time vs. rush-hour}: Here, we choose some trips that start early in the morning--3~am to 5~am--and some that start during the NYC rush hour--4~pm to 6~pm.
    \item \emph{Low vs. high temperature}: In this scenario, we compare trips with a relatively low temperature--trips that are in the 0.25 quartile and not in the 0.1 decile--with trips that took place at a high temperature--trips that are in the 0.75 quartile and not in the 0.9 decile.
    \item \emph{Low vs. high distance}: We select trips with a relatively high/low distance between pickup and dropoff locations--we use the same boundaries as for SC3 for the feature \emph{Haversine distance} to select the trips.
\end{enumerate}

\section{Results and Discussion}

\subsection{Models for Estimating the Time of Arrival}
\label{sec:ModelsForETA}
We take the three \ac{ETA} models proposed by \textcite{schleibaum.2022} as our first-level models as well as their hyperparameters, which have been chosen via Bayesian optimization. The first model is based on \ac{RF} (\emph{L1-RF}) with 300 trees and a maximum tree depth of 89; the number of maximal features per node is chosen automatically and the minimum number of samples per leaf and split are set to four. The second model is based on XGBoost (\emph{L1-XGBoost}) and also consists of 300 trees, but has a maximum tree depth of eleven; the minimum number of instances required in a child is set to seven, the subsample ratio of the training data per tree to one, the minimum loss reduction required for making a further partition on a child to zero, and the subsample ratio of features for a tree to one. The third model is based on a \ac{NN} (\emph{L1-FCNN}) with four hidden layers and 300, 150, 50, and 25 corresponding neurons. Similarly to \textcite{schleibaum.2022}, we set the batch size to 128, the learning rate to 0.001, train the network for 25 epochs, and select the best model along the training to minimize overfitting.

Besides the first-level models, we propose four second-level models or ensembles. Because we use all three first-level models for each ensemble, all four ensembles are heterogeneous. The first second-level model is a relatively simple one based on an \ac{MLR} referred to as \emph{L2-MLR}. The second one is based on \ac{RF} (\emph{L2-RF}) with 100 trees in the forest; for the third, XGBoost-based model or \emph{L2-XGBoost}, we chose the same number of trees. For both--\emph{L2-RF} and \emph{L2-XGBoost}--we do not train the hyperparameters as these methods usually achieve a high prediction precision without any hyperparameter tuning. The fourth ensemble (\emph{L2-FCNN}) combines the output of the first-level models via a \ac{NN} with two fully connected hidden layers--50 and 25 corresponding neurons--and otherwise similar hyperparameters to the \emph{L1-FCNN}.

Table~\ref{tab:ComparisonOfMLModels} shows that for the New York City data set the \acs{MAE} or average prediction error in seconds per trip is around 178 seconds for the L1-\ac{NN} and a couple of seconds higher for the L1-\ac{RF} and L1-XGBoost. The results for the other evaluation metrics--the \acs{MRE} and \acs{MAPE}--are similar and put the prediction error in perspective to the trip duration. Regarding the New York City data set and the second-level models, all models are able to outperform the first-level models with regard to \acs{MAE} and \acs{MRE}. Only the L2-\acs{NN} is able to outperform all first-level models in all evaluation metrics with an \acs{MAE} of 169 seconds or an \acs{MRE} of around 20 percent. Interestingly, the L2-\acs{MLR} achieves a remarkable prediction precision that is better than that of the L2-\acs{RF} and similar to that of the L2-XGBoost. For the models trained and tested on the Washington DC data set, we observe that on the first level, the L1-\ac{NN} with an \ac{MAE} of around 169 seconds is able to outperform the L1-\ac{RF} by 10 seconds and the L1-XGBoost by 20 seconds. Regarding second-level models, we observe that except for the L2-\acs{RF} all models achieve a remarkable prediction precision. In contrast to the models trained and tested on the New York City data set, none of the second-level models is able to outperform the \ac{MAPE} achieved by the L1-\ac{NN}.

As shown in the last three lines of Table~\ref{tab:ComparisonOfMLModels}, we outperform the baseline approaches by at least 15 seconds on the New York City data set and by at least 6 seconds on the Washington DC data set when considering the \ac{MAE}. The results are similar for the other two evaluation metrics.
 
\begin{table}[ht]
	\caption{Comparison of our \ac{ETA} models of the first and second level based on different evaluation metrics}
	\begin{tabular*}{\hsize}{@{\extracolsep{\fill}}llcccccc@{}}
		\toprule
		            & Data set               & \multicolumn{3}{c}{New York City}                 & \multicolumn{3}{c}{Washington DC}              \\
		\midrule
		            & Evaluation metric     & \acs{MAE} [seconds]   & \acs{MRE} &   \acs{MAPE}  & \acs{MAE} [seconds]   & \acs{MRE} & \acs{MAPE} \\
		\midrule
		Level 1     & L1-\acs{RF}           & 180.694               & 0.2158    &   27.8689     & 179.5912 & 0.2373 & 30.1512 \\
		            & L1-XGBoost            & 183.4192              & 0.219     &   27.1137     & 190.2613 & 0.2514 & 30.4033 \\
		            & L1-\acs{NN}           & 178.2321              & 0.2129    &   23.7561     & 169.8152 & 0.2244 & \textbf{24.372} \\
		\midrule
		Level 2     & L2-\acs{MLR}          & 172.2439              & 0.2057    &   25.2758     & 171.178  & 0.2262 & 27.1985 \\
		            & L2-\acs{RF}           & 183.2319              & 0.2188    &   26.9828     & 183.7377 & 0.2428 & 29.5762 \\
		            & L2-XGBoost            & 173.6526              & 0.2074    &   25.3077     & 172.7287 & 0.2283 & 27.5419 \\
		            & L2-\acs{NN}*          & \textbf{169.4285} & \textbf{0.2023} & \textbf{22.9121} & \textbf{167.9959} & \textbf{0.222} & 24.6133 \\
        \midrule
		Baselines   & \cite{jindal.2018}    & 185.9265              & 0.2256    &   23.8429     & 181.1275 & 0.2374 & 27.4261 \\
		            & \cite{dearaujo.2019}  & 201.5998              & 0.2455    &   28.1508     & 203.8581 & 0.2673 & 35.4898 \\
		            & \cite{haliem.2021a}   & 185.3999              & 0.2257    &   28.3598     & 174.3907 & 0.2286 & 25.7570 \\
		\bottomrule
		\multicolumn{8}{l}{*This prediction precision is better than the one presented by \textcite{schleibaum.2022}}
	\end{tabular*}
	\label{tab:ComparisonOfMLModels}
\end{table}

\subsection{Explaining the First-Level Estimated Time of Arrival Models}
\label{sec:ExplainingFirstLevelModelsForETA}

\begin{figure}
    \centering
    \begin{subfigure}[b]{0.49\textwidth}
        \centering
        \includegraphics[width=\textwidth]{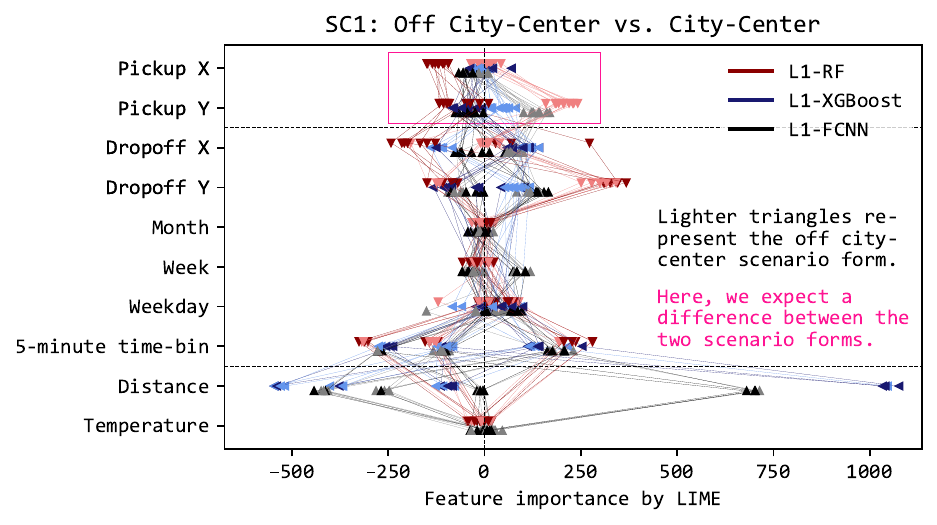}
        \caption{}
    \end{subfigure}
    \hfill
    \begin{subfigure}[b]{0.49\textwidth}
        \centering
        \includegraphics[width=\textwidth]{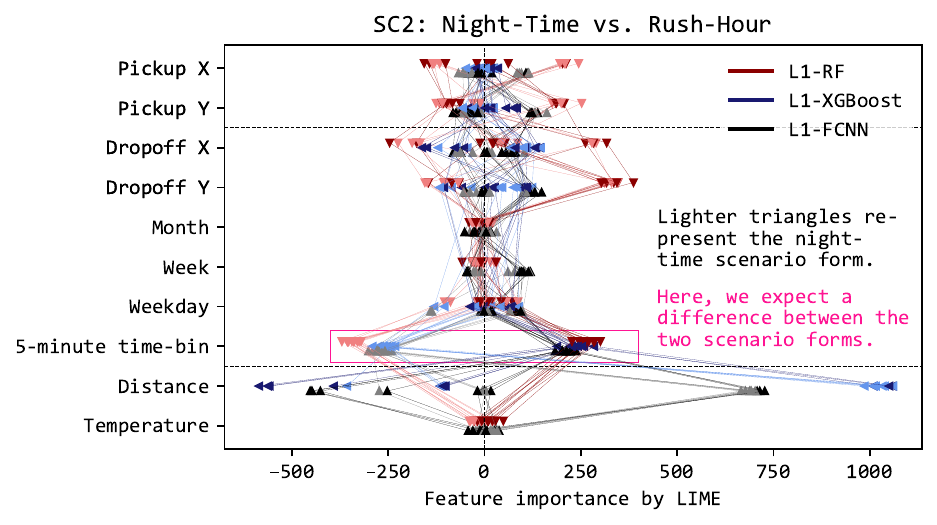}
        \caption{}
    \end{subfigure}
    \hfill
    \begin{subfigure}[b]{0.49\textwidth}
        \centering
        \includegraphics[width=\textwidth]{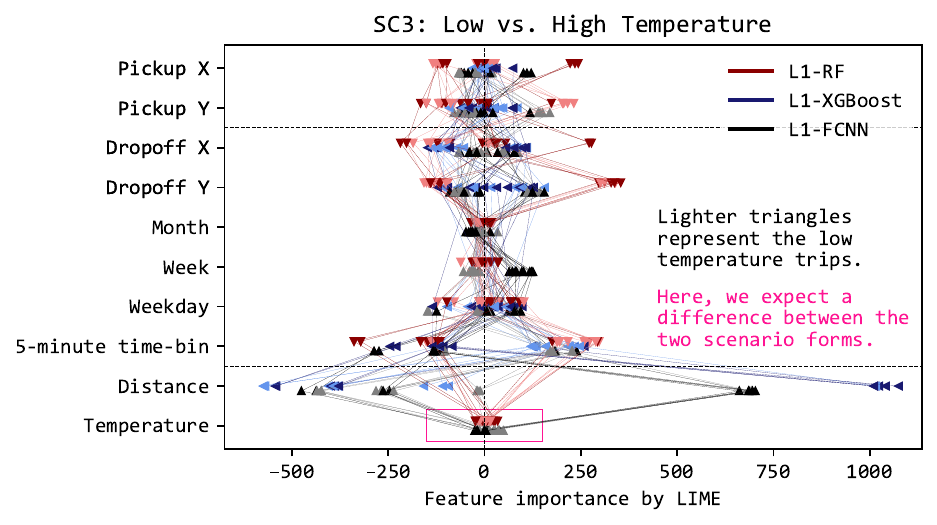}
        \caption{}
    \end{subfigure}
    \hfill
    \begin{subfigure}[b]{0.49\textwidth}
        \centering
        \includegraphics[width=\textwidth]{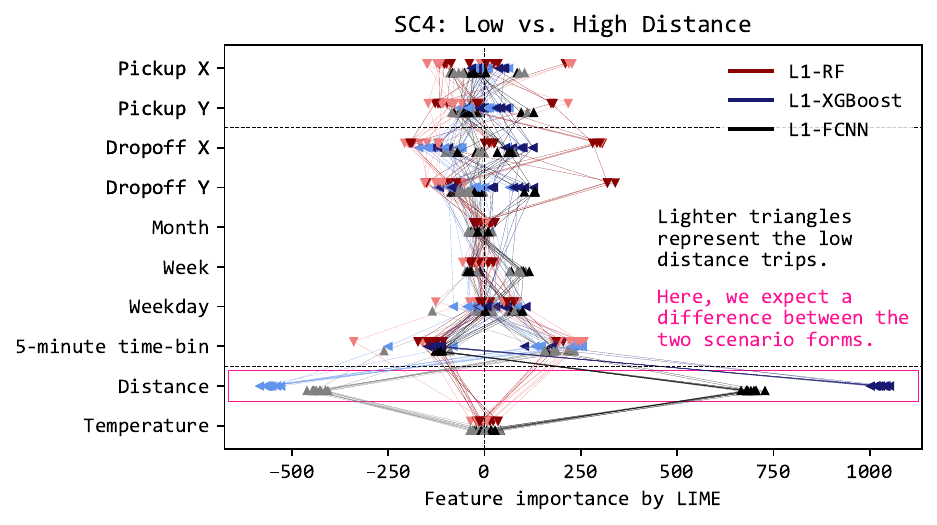}
        \caption{}
    \end{subfigure}
    \caption{Local feature importance via \ac{LIME} per feature of the samples in the scenarios for the first-level models; each plot refers to one scenario, (a) for SC1, (b) for SC2, (c) for SC3, and (d) for SC4; the ten trips with an expected lower influence are marked with lighter triangles--the ones with an expected higher influence with less light triangles; each line connects the feature importances for one trip along the various features used by the corresponding model.}
    \label{fig:Lev1ScenariosLIME}
\end{figure}

In Figure~\ref{fig:Lev1ScenariosLIME}, we visualize the explanations generated by \ac{LIME} per scenario, its characteristics, feature, and \ac{ETA} model. Each triangle represents the importance of a feature for one trip from a scenario--the lighter triangles are from the first or `lower` characteristic of the scenario. The lines connect the importance of the features for one sample or trip. Concrete values can be interpreted as follows: for instance, the left-most triangle in SC1--(a) of Figure~\ref{fig:Lev1ScenariosLIME}--has a value of around $-575$. This refers to a relatively strong negative influence of the concrete distance value on the corresponding estimated duration of the trip. This most likely refers to a trip with a small distance between the pickup and dropoff locations.  

In SC2--top right--and SC4--bottom right--the two characteristics of each scenario are visually separated for the features that constitute the scenario--the 5-minute time-bin for SC2 and the distance for SC4. As expected, the other features are not separable because they are more or less randomly distributed over the space of each feature--not completely random, as some features might slightly correlate with the feature of interest. Moreover, the separation is in the correct order, which means that the `lower` characteristic also has a lower importance of features than the `higher` characteristic of the scenario. Therefore, the \ac{ETA} models appear to have properly learned the expected behavior in these scenarios. Even though for the majority of trips in SC1 the difference between the pickup off city-center and in city-center is learned, some trips of the 'lower' and 'higher' characteristics interfere. For instance, for \emph{pickup X}, the \emph{L1-\acs{NN}} feature importances of both scenario characteristics overlap. As regards SC3, we observe that the reported feature importances for the temperature are low or close to zero. While this could indicate that the \ac{ETA} models have not learned the underlying pattern, similarly to \textcite{schleibaum.2022}, we argue that the overall feature contribution of the weather or temperature is low.

\begin{figure}
    \centering
    \begin{subfigure}[b]{0.49\textwidth}
        \centering
        \includegraphics[width=\textwidth]{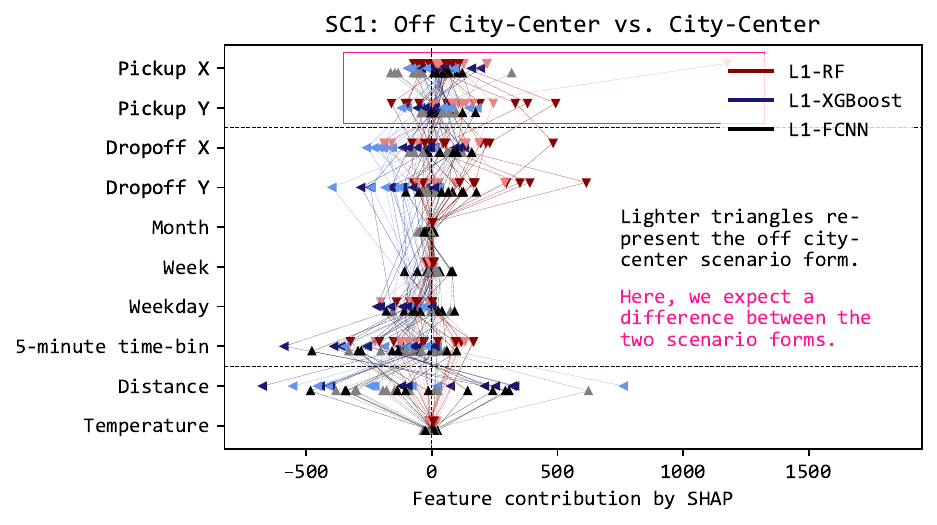}
        \caption{}
    \end{subfigure}
    \hfill
    \begin{subfigure}[b]{0.49\textwidth}
        \centering
        \includegraphics[width=\textwidth]{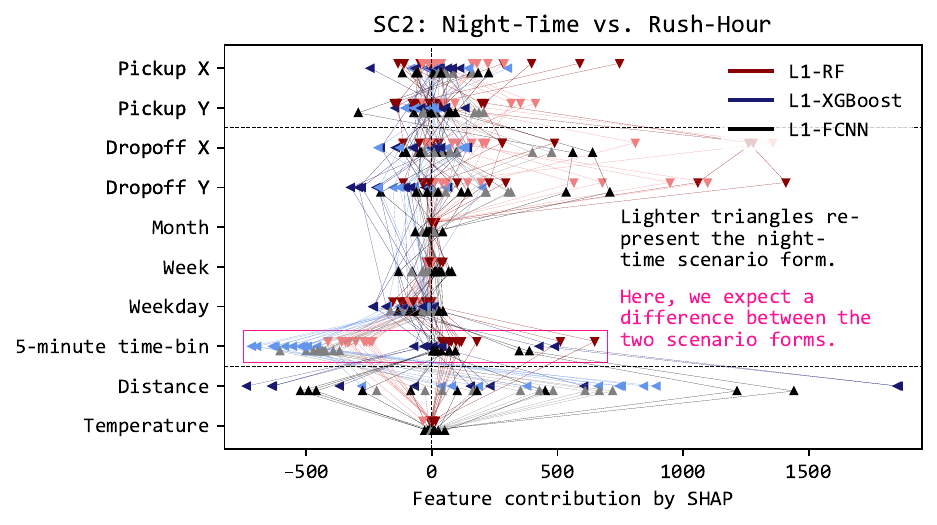}
        \caption{}
    \end{subfigure}
    \hfill
    \begin{subfigure}[b]{0.49\textwidth}
        \centering
        \includegraphics[width=\textwidth]{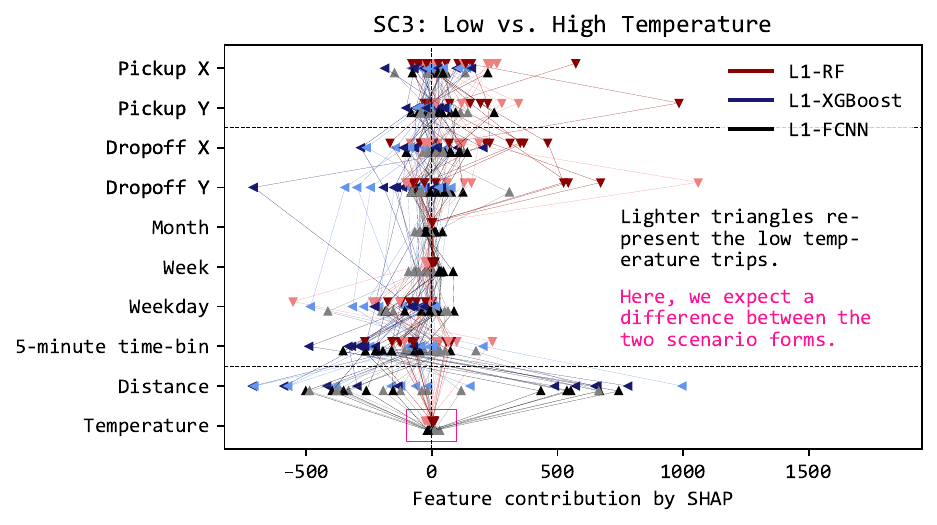}
        \caption{}
    \end{subfigure}
    \hfill
    \begin{subfigure}[b]{0.49\textwidth}
        \centering
        \includegraphics[width=\textwidth]{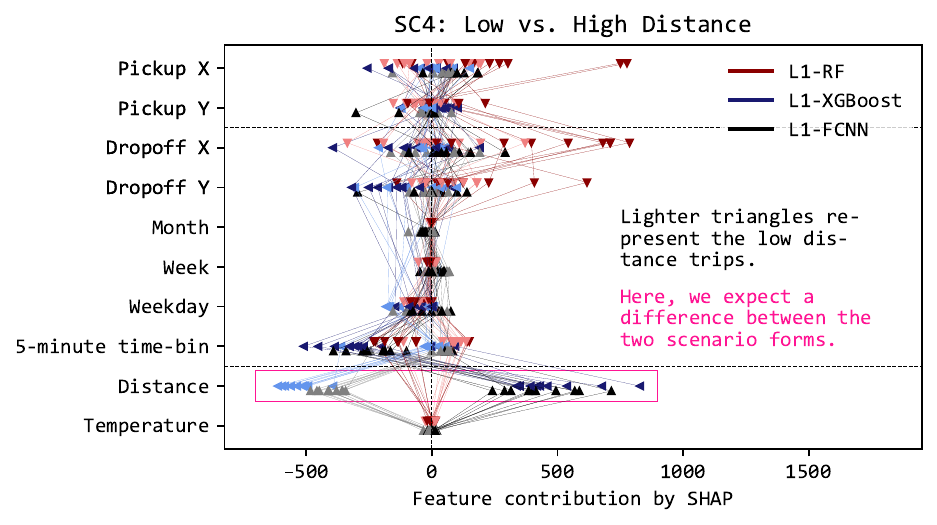}
        \caption{}
    \end{subfigure}
    \caption{Explanations via \ac{SHAP} per feature, sample, and scenario for the first-level models; each subfigure refers to one scenario, (a) for SC1, (b) for SC2, (c) for SC3, and (d) for SC4; the ten trips with an expected lower influence are marked with lighter triangles--the ones with an expected higher influence with fewer light triangles.}
    \label{fig:Lev1ScenariosSHAP}
\end{figure}

While the concrete values or feature contributions generated via \ac{SHAP} differ from the feature importances of \ac{LIME}, we observe similar results in Figure~\ref{fig:Lev1ScenariosSHAP}: For SC2 and SC4, the two characteristics of the scenarios are visually separated only by the feature of interest; the separation for SC1 and SC3 is not clear.  

\subsection{Explaining the Ensemble Model for Estimated Time of Arrival}
\label{sec:ExplainingEnsembleModelsForETA}
In the following, we describe three relatively simple but novel methods to join the explanations from the first and second level. To be able to compare the outputs for all three joining methods and samples in a better way, we first normalize the second-level explanation per sample so that they sum up to one. The joining methods (JMs) are: 
\begin{enumerate}[label=(JM\arabic* ]
    \item - \emph{Adding a Dimension}) Here, we simply output both, the first and second-level explanations--which is meant by 'additional dimension'--simultaneously. Therefore, we weigh the feature contribution or importance of a feature at the first level with the contribution or importance of the prediction at the first level at the second level. Consequently, we join the first and second-level explanations without losing any information.  
    \item - \emph{Basic Join of the Contributions}) To determine the contribution or importance of a feature, we compute the dot product between the vector that contains the contribution or importance of that feature for each first-level model and the vector that contains the contribution or importance of each first-level model on the second level; the product is then the joint contribution or importance of that feature for a given sample.  
    \item - \emph{Diversifying the Contributions}) Here, we use the result from JM2 as a basis and define a threshold $\alpha$--which is the mean value of the distribution or influence of each first-level model on the second level or e.g. $1/3$ with our three first-level models. Next, every value below that threshold is reduced by a value $\beta$ to be increased by the collected value in the next step. If the values cannot be reduced by $\beta$, because otherwise they would become negative, only the difference to zero is used and redistributed to the values above $\alpha$. Thus, the second-level influence is diversified. In the following, we set $\beta$ to $0.5$ or relatively high, as the number of first-level models is only three.
\end{enumerate}

\begin{figure}[ht]
    \centerline{\includegraphics[width=\textwidth]{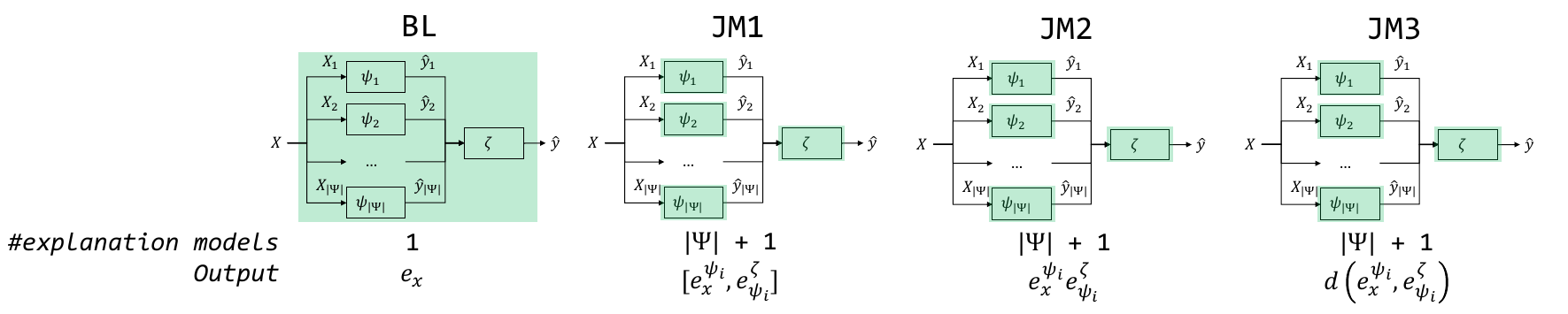}}
    \caption{Overview of our three joining methods (JM1, JM2, JM3) compared to the baseline (BL). Those models to which an explanation method is applied are highlighted in green, showing that the number of explanation models is higher (providing more insights) in our proposed explanation methods. Further, we show the output--$e$ refers to an explanation and $d()$ to the diversifying function described previously.}
    \label{fig:XAIMethodIllustration}
\end{figure}

All three joining methods are compared to a \emph{\ac{BL}} method. This method generates explanations by explaining a function that wraps the whole ensemble. Within this function, features that are an alternative representation of other features, like the X-index of a 50-meter square grid, are also generated within that function from the corresponding base feature, i.e. the latitude of the pickup location. In Figure~\ref{fig:XAIMethodIllustration}, we show an overview of our three joining methods compared to the baseline. 

\begin{figure}
    \centering
    \begin{subfigure}[b]{0.49\textwidth}
        \centering
        \includegraphics[width=\textwidth]{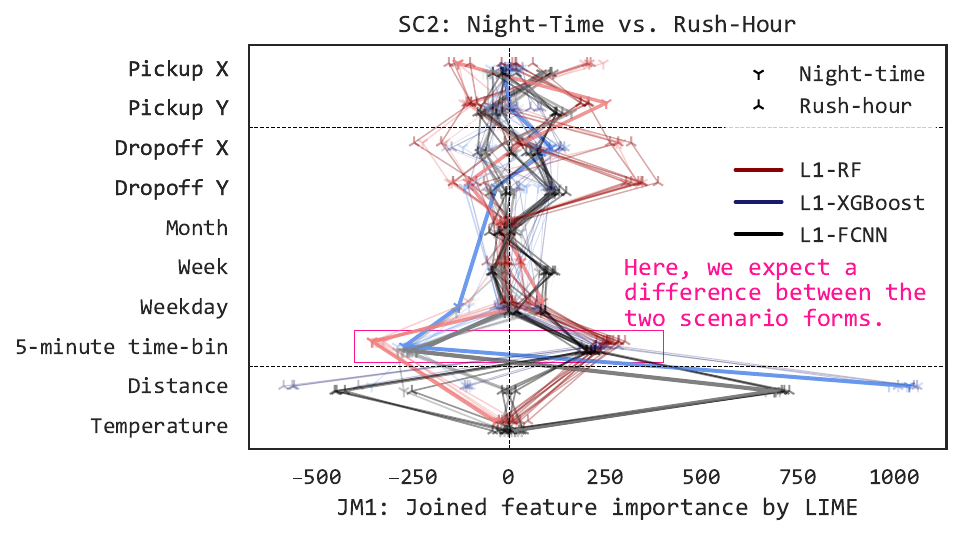}
        \caption{}
    \end{subfigure}
    \hfill
    \begin{subfigure}[b]{0.49\textwidth}
        \centering
        \includegraphics[width=\textwidth]{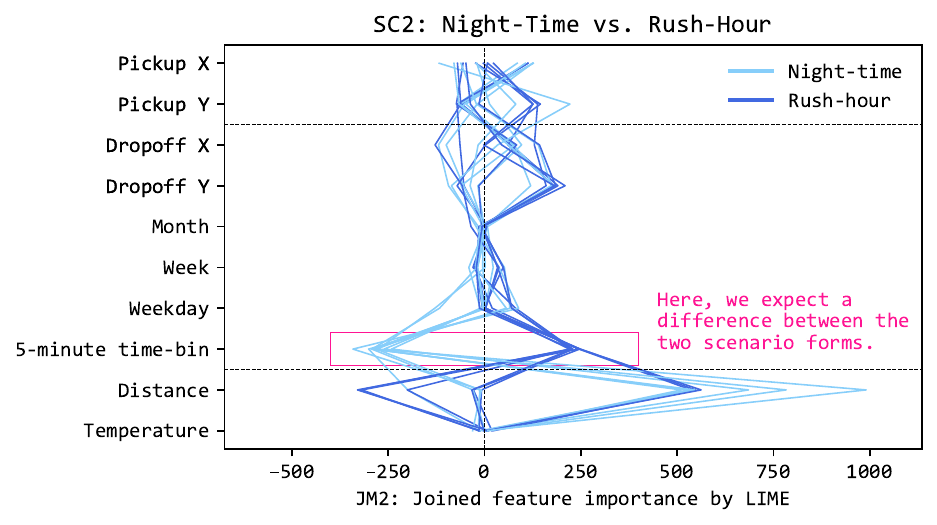}
        \caption{}
    \end{subfigure}
    \hfill
    \begin{subfigure}[b]{0.49\textwidth}
        \centering
        \includegraphics[width=\textwidth]{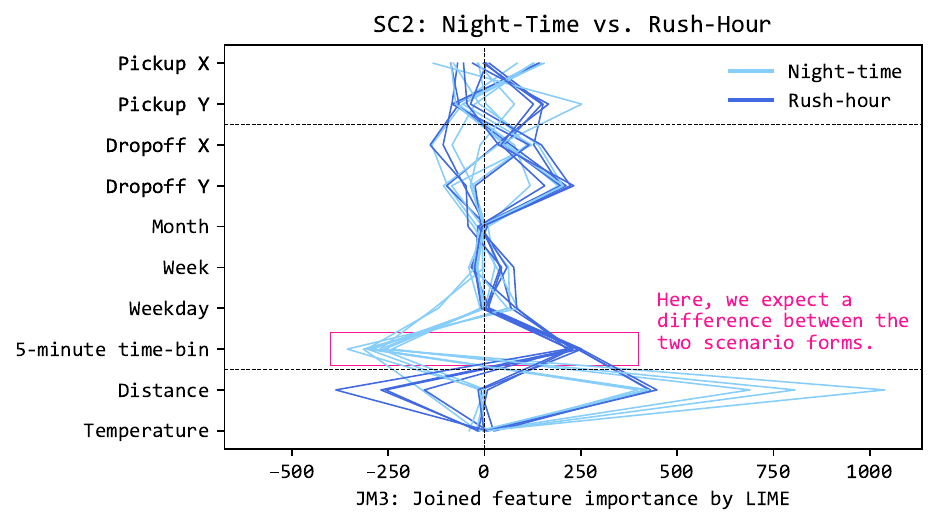}
        \caption{}
    \end{subfigure}
    \hfill
    \begin{subfigure}[b]{0.49\textwidth}
        \centering
        \includegraphics[width=\textwidth]{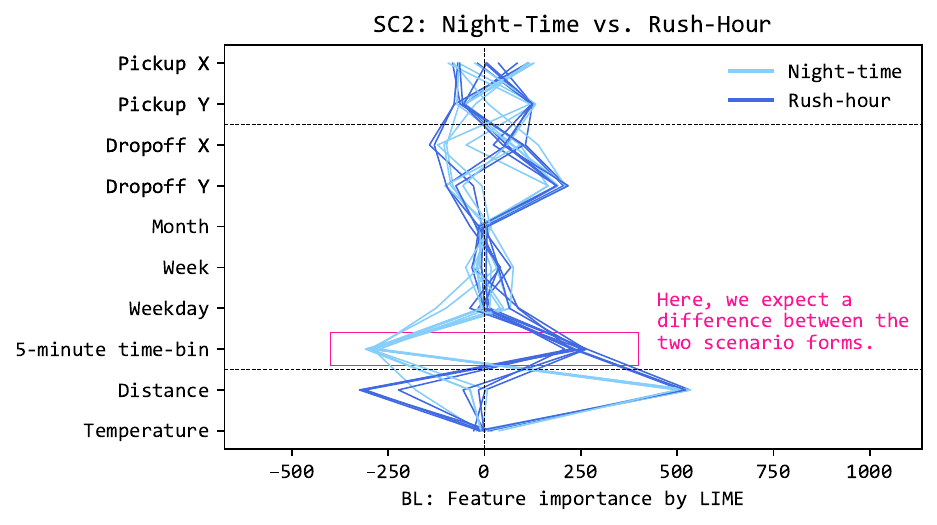}
        \caption{}
    \end{subfigure}
    \caption{Joint local feature importance via \ac{LIME} for each feature of the samples in the second scenario (SC2--night-time vs. rush-hour) for the joining methods JM1 (a), JM2 (b), and JM3 (c) compared to the \ac{BL} (d). Each line connects the feature importances for one trip and in (a) the line width corresponds to the influence on the second level.}
    \label{fig:Lev2Scenario2LIMEv3}
\end{figure}

In Figure~\ref{fig:Lev2Scenario2LIMEv3}, we show the feature importance for \ac{LIME} joined via all three proposed joining methods as well as for the \ac{BL}. Similar to previous findings, in each graphic of Figure~\ref{fig:Lev2Scenario2LIMEv3}, for each trip and method a line--or three for JM1--is shown, this time without triangles. As we only want to demonstrate the joining methods, we omit all scenarios except for SC2 here, but we show the corresponding graphics for SC1, SC3, and SC4 in Figure~\ref{fig:Lev2Scenario134LIMEv3} in the Supplementary Material.

In Figure~\ref{fig:Lev2Scenario2LIMEv3}, the difference between JM1 and \ac{BL}/JM2/JM3 is obvious: JM1 shows much more information, including the proof that all three first-level models are used by the \emph{L2-FCNN}. Even though the two scenario characteristics have the expected difference at the 5-minute time-bin, verifying the difference among the first-level models is hard for JM1. Regarding JM2, we observe a relatively high difference to the \ac{BL} joining method as for instance visible in the feature importances of the 5-minute time-bin or the distance of the night-time characteristic of the scenario. As expected, the JM3 makes the smaller feature importance values smaller and the larger ones larger, thereby diversifying the feature importances along all features slightly.

\begin{figure}
    \centering
    \begin{subfigure}[b]{0.49\textwidth}
        \centering
        \includegraphics[width=\textwidth]{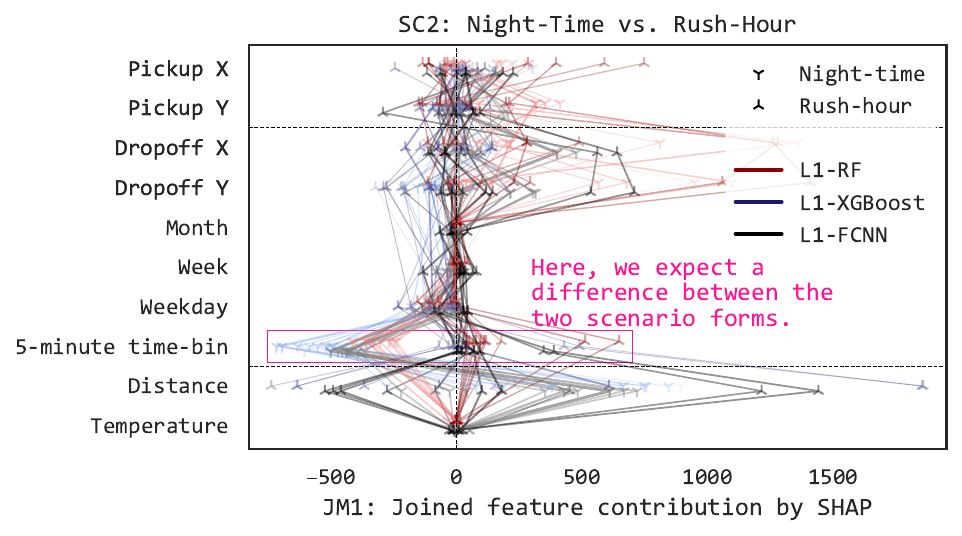}
        \caption{}
    \end{subfigure}
    \hfill
    \begin{subfigure}[b]{0.49\textwidth}
        \centering
        \includegraphics[width=\textwidth]{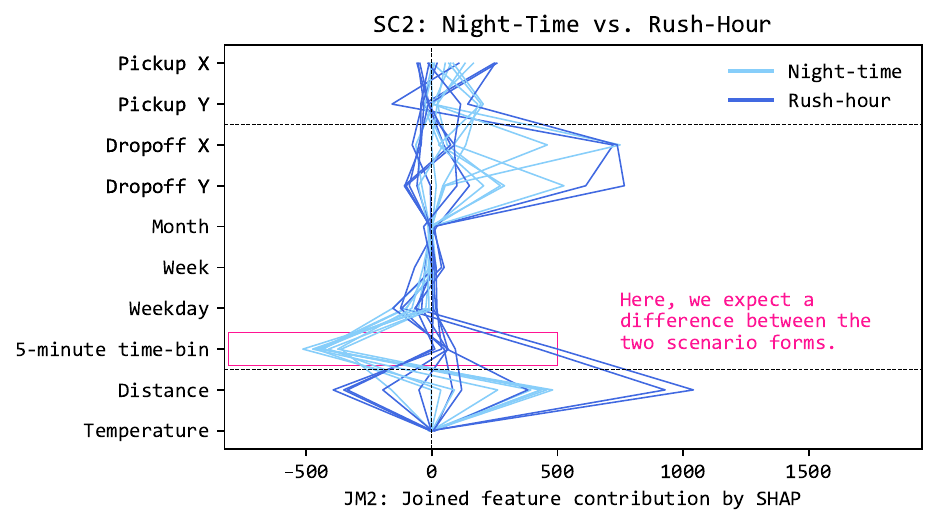}
        \caption{}
    \end{subfigure}
    \hfill
    \begin{subfigure}[b]{0.49\textwidth}
        \centering
        \includegraphics[width=\textwidth]{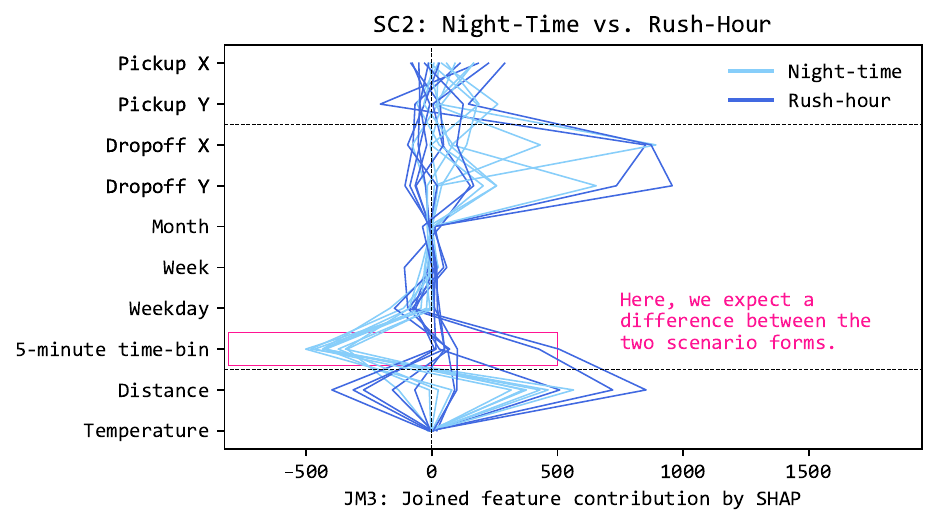}
        \caption{}
    \end{subfigure}
    \hfill
    \begin{subfigure}[b]{0.49\textwidth}
        \centering
        \includegraphics[width=\textwidth]{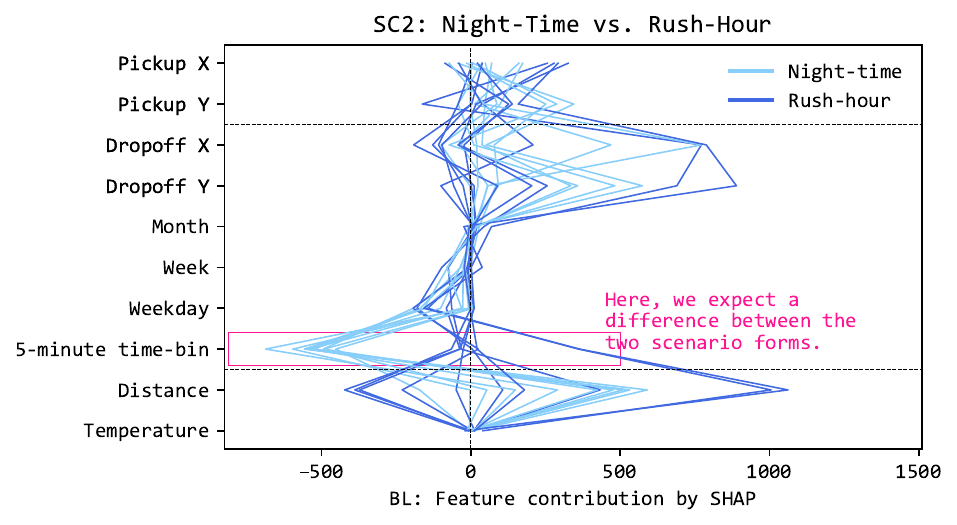}
        \caption{}
    \end{subfigure}
    \hfill
    \caption{Joint local feature importance via \ac{SHAP} for each feature of the samples in the second scenario (SC2--night-time vs. rush-hour) for the joining methods JM1 (a), JM2 (b), and JM3 (c) compared to the \ac{BL} (d). Each line connects the feature importances for one trip and in (a) the line width corresponds to the influence on the second level.}
    \label{fig:Lev2Scenario2SHAPv3}
\end{figure}

When applying the joining methods to the \ac{SHAP} values for the same scenario, as shown in Figure~\ref{fig:Lev2Scenario2SHAPv3}, we observe similar results. While the difference between the night-time and rush-hour characteristic of SC2 is visible for all joining methods, this time JM2 and JM3 in general reduce the feature importances. This is in contrast to the explanations generated by \ac{LIME}. 

\begin{figure*}
    \centering
    \includegraphics[width=\textwidth]{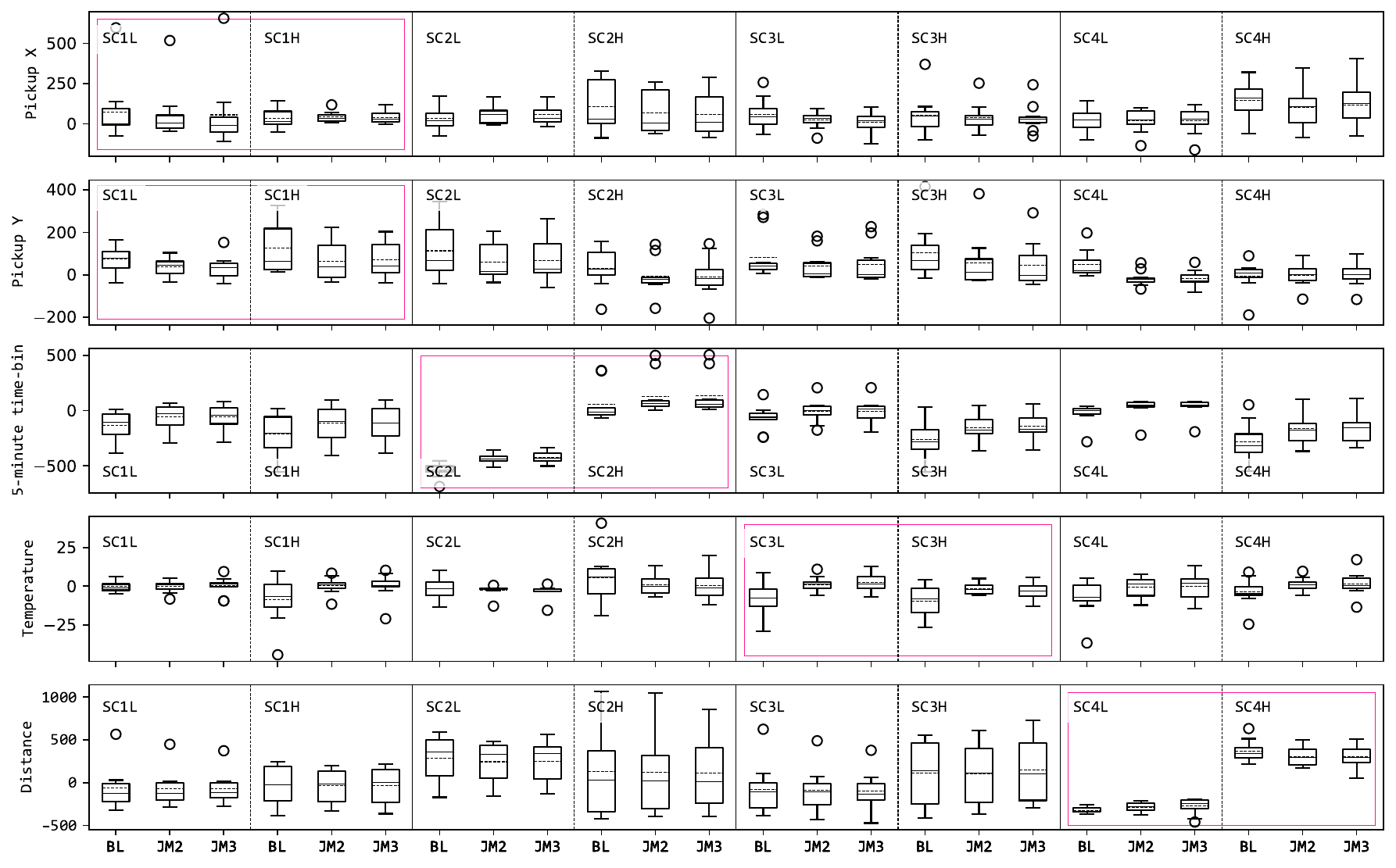}
    \caption{Box plot per feature--those affected by the scenarios like the 5-minute time-bin for SC2--and joining method compared to the \ac{BL} for each scenario in its lower (e.g. \emph{SC1L}) and its higher (e.g. \emph{SC1H}) characteristic; the dashed lines in the boxes of each box plot are the mean values and the pink rectangles mark the features of the scenarios.}
    \label{fig:SHAPSCFeaturesBoxplots}
\end{figure*}

In Figure~\ref{fig:SHAPSCFeaturesBoxplots}, we visualize the Shapley values for the features used to build the scenarios via the joining methods JM2 and JM3 per scenario and their two opposing characteristics--`lower' and `higher'--to further investigate the differences to the BL; JM1 is omitted in the figure as it is hard to compare in the visualized regard. As expected, the Shapley values generated via JM2 and JM3 do not vary much compared to the \ac{BL}; like for the 5-minute time-bin and SC2H, JM2 and JM3 slightly change the Shapley values in the positive direction. For the distance and SC4L, the Shapley values are moved in the opposite direction. In general, the difference expected in the scenarios gets slightly smaller, but it is still clearly shown. A similar figure for LIME can be found in Figure~\ref{fig:LIMESCFeaturesBoxplots} the Supplementary Material.

\subsection{Discussion}
\label{sec:Discussion}

\paragraph{Ensemble for \ac{ETA}.} We developed multiple alternatives to combine the outputs of the \ac{RF}, XGBoost and \ac{NN} models via another model. Even though several second-level models achieved a high prediction precision on the New York City data set, only an \ac{NN}-based one was able to outperform our previous models from \textcite{schleibaum.2022} in all evaluation metrics. Interestingly, for the Washington DC data set the results were not that clear: while the \ac{NN}-based ensemble performed better than the first-level \ac{NN}-based model as regards the \ac{MAE} and \ac{MRE}, for the \ac{MAPE} the observed pattern is the opposite. We believe that this is caused by three reasons. First, feature selection and hyperparameter tuning were done for the New York City data set and consequently not optimal for the Washington DC data set. Second, the Washington DC data set with around 650K trips used is much smaller than the New York City one with 1.25M trips, causing the second-level models to have much fewer data to be trained on. Third, the gap between the performance of the three first-level models is much closer for the models trained on the New York City data set compared to those first-level models trained on the Washington DC data set. Therefore, we assume that a better-performing XGBoost model or excluding it from the ensemble could further improve the prediction precision. As we already outperformed the approaches of \cite{haliem.2021a,dearaujo.2019,jindal.2018} in our previous work--\cite{schleibaum.2022}-- we consider the usage of a stacked heterogeneous ensemble as an effective method to increase the prediction precision for static route-free \ac{ETA}. With the data set considered, we reduced the \ac{MAE} by nine seconds to around 169 seconds per trip on average; both \ac{MRE} and \ac{MAPE} were reduced by around one percentage point.

\paragraph{Explaining First-Level \ac{ETA} Models.} We applied the two model-agnostic \ac{XAI} methods \ac{LIME} and \ac{SHAP} to evaluate and explain our first-level \ac{ETA} models post-hoc and locally. In SC2 and SC4, we could show that all three models learned the expected behavior. For SC3, all \ac{ETA} models that include the temperature consistently learned a low influence of the temperature on the \ac{ETA}. As described previously, this is most likely caused by the low influence of weather-related data in general rather than a pattern that is not properly learned by our \ac{ETA} models. Even though \textcite{schleibaum.2022} showed the positive influence of including the feature in the models on the prediction precision, as their general influence is low, the explanation or feature importance value assigned is not very meaningful. In case someone focuses on explainable \ac{ETA} models, removing the temperature or the month feature might be worth considering. Regarding SC1, we could show that information like the pickup location, which is encoded into multiple and, therefore, correlating features, is difficult to explain by \ac{LIME} and \ac{SHAP}. We observed that the explanations produced by \ac{LIME} are more separated in our scenarios compared to those of \ac{SHAP}. As the focus of our work is not to compare \ac{LIME} and \ac{SHAP}, we refer the interested reader to the work of \textcite{belle.2021}; but in general, \ac{LIME} has a relatively low runtime and \ac{SHAP} has the advantage of producing additive explanations. 

\paragraph{Joint Explanation of Ensembles for \ac{ETA}.} We presented three relatively simple methods for joining the first- and second-level explanations of an ensemble to generate a joint explanation. The main advantage and at the same time drawback of joining method JM1 (\emph{Adding a Dimension}) is that more information or all explanations are shown. When--as we did--multiple trips are shown in one graphic, we assume that it is harder to understand, but on the other hand, especially when only one trip is visualized, this provides additional insights not provided by JM2 and JM3. For instance, it might be interesting to see if different first-level models disagree on a feature's importance for a specific sample, how strong the influence of each of the models is, and if some relation was not learned correctly. For such a case, a smarter choice of colors or an alternative to the used line plots could improve the understandability. However, with respect to larger ensembles or those that have more or many first-level models, the less dense explanation created by JM1 might be confusing or not understandable anymore.

As regards joining method JM2 (\emph{Basic Join of the Contributions}), we observed an unexpectedly high difference to the \ac{BL} method. We believe that this difference is at least partly caused by correlated features such as the pickup latitude and longitude. Nevertheless, the general direction of the feature importances is similar. Even though the \ac{XAI} methods might not be built for correlated features, especially in stacked ensembles and practice, correlated features exist. Interestingly, when considering \ac{LIME}, the larger values are made larger; for \ac{SHAP}, the effect is the opposite. For JM3 (\emph{Diversifying the Contributions}), we observed the same but slightly stronger effect. In contrast to JM1, JM2, and the \ac{BL}, JM3 has a hyperparameter that has to be chosen by the user, which makes this method more complicated to apply. 

Interestingly, none of the related work--\cite{silva.2019, kallipolitis.2021, rozemberczki.2021}--has used the \ac{BL} method to generate an explanation or compare their explanation to it. As we did not find other literature regarding explaining ensembles, we assume that our work presents three novel joining methods. While the general concept that we applied to create a joint explanation of a stacked ensemble with two levels that performs a regression is relatively simple, the proposed concept is neither specific to the underlying \ac{XAI} method nor to the regression models. It could even be applied to the probabilities generated by classification models. Additionally, the concept does not depend on the number of first-level models and can, as we did, be applied to first-level models that only share a part of their input features. Also, the joining methods are model-agnostic and a combination of different \ac{XAI} methods is possible. When, for instance, considering one or multiple complex models on the first level, explaining them with an \ac{XAI} method that has a faster inference time, and combining that on the second level with an \ac{XAI} method like \ac{SHAP}, is possible. 

\paragraph{Limitations and Future Work.} As argued before, we applied relatively moderate criteria for identifying outliers before training the various \ac{ETA} models. We did this to make the comparison to non-reproduced papers fairer. However, we expect that we could further increase the prediction precision of the composed ensemble model. Another option to potentially achieve a higher prediction precision is to include other \ac{ETA} models into the ensemble as additional first-level models, for instance, \cite{haliem.2021a, dearaujo.2019, kankanamge.2019, li.2018}.

While the evaluation of the performance of \ac{ETA} models is relatively straightforward, the evaluation of explanations is not; especially, determining the influence of the slight differences between our joining methods is affected by this problem. Moreover, we considered only four self-chosen scenarios to demonstrate and evaluate the generated explanations; many more scenarios like the ones that combine features--for instance, pickup at the city center during rush hour--might be interesting and valuable for evaluation. While we focused on generating explanations in a general way so that others can adapt and build upon our work, correlated features or information that span over multiple features like the pickup location could be explained better when the features are explained jointly or the x and y values of the pickup location and their influence on the \ac{ETA} are visualized on a map. 
% Drawback of combination--so far only importance but not direction of importance

Regarding explanations, future work will look into ways to explain information that spans over multiple features. Another option to extend this work is to use other \ac{XAI} methods or use different ones for different models to generate more accurate explanations per model type. The latter could also be used to generate explanations relatively fast, for instance by using \ac{LIME} on first-level models with a greater feature space and \ac{SHAP} on the second-level models with smaller feature space. Moreover, the explanations generated here are vectors of values and, therefore, still hard to understand by affected users like taxi drivers or passengers. The explanations could be translated into more human-friendly ones, for instance, by linking an explanation of the locations` influence on the \ac{ETA} to points of interest like the main train station that is close to the dropoff location and thus possibly increasing the \ac{ETA} for an upcoming trip. Moreover, our explanation could be enhanced from route-free to route-based \ac{ETA} as such approaches are more likely to be used by taxi drivers and passengers thanks to their increased prediction precision. Additionally, using the generated explanations might be beneficial not only for users of an \ac{ETA} model but also for the designers of such models. Based on the explanations, some first-level models or features used in a model might be excluded--leading to a smaller and more precise \ac{ETA} model.

\section{Conclusions}
\label{sec:Conclusion}
On-demand transportation modes, such as ride-sharing or ride-hailing, are key building blocks of sustainable passenger transportation. Estimating the time of arrival of vehicles (taxis) in ride-sharing or ride-hailing is relevant for the comparison and computation of schedules, and provides important information to drivers and passengers. In this paper, we investigate how the prediction precision of \ac{ETA} algorithms can be improved by combining multiple \ac{ML} models into a stacked heterogeneous ensemble--which, on its own, is novel and has been shown to outperform state-of-the-art static route-free \ac{ETA} methods on two data sets. Furthermore, to enable the explainability and transparency of the stacked model, we proposed \ac{XAI} methods for explaining the first-level models of the ensemble, as well as three novel methods for joining the first and second-level explanations of the ensemble model. To demonstrate the feasibility and benefit of our approach, we use a taxi trip data set collected in New York City to evaluate our explanations against a baseline model that wraps the whole ensemble in one function. Based on the limitations, more tuning of the ensemble models and the inclusion of other \ac{ETA} models from the related work is promising. In addition, we want to explore the explanation of correlated features and the combination of different \ac{XAI} methods to explain ensembles.

\section*{Data Availability} 
As described in Section~\ref{sec:Dataset}, we use the New York City Yellow taxi trip data from 2015 and 2016, which is publicly available, and the Washington DC taxi trip data recorded in 2017 to support the findings of this study. Links to the data sets are included in the references--\cite{newyork.2022} and \cite{kaggle.2019}. Methods used to enhance the data sets by the additional features considered throughout this paper are provided in our code repository--see \cite{codeeta.2022}.

\section*{Conflicts of Interest}
The authors declare that there is no conflict of interest with respect to the publication of this article.

\section*{Funding Statement}
This work was supported by the Deutsche Forschungsgemeinschaft under grant 227198829/GRK1931. The SocialCars Research Training Group focuses on future mobility concepts through cooperative approaches.

\section*{Acknowledgments} 
A preprint has previously been published \cite{scibaum.2022}. Additionally, we thank Steven Minich, Helene Nicolai, and Julian Teusch for providing helpful feedback--especially regarding the explanation part of the paper. 

\newpage
\section*{Supplementary Material}
\label{sec:SupplementaryMaterial}

\begin{figure}[ht]
\centering
\begin{subfigure}[b]{\textwidth}
    \includegraphics[width=0.246\linewidth]{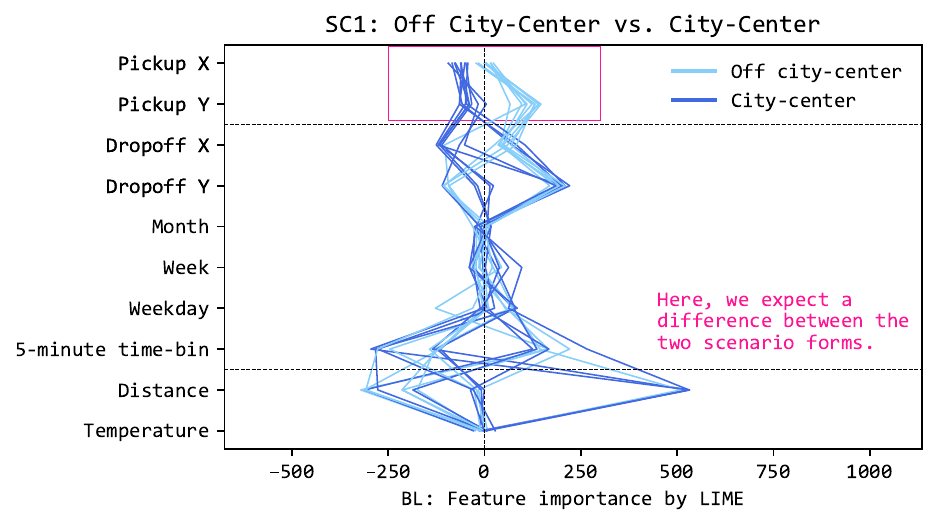}
    \hfill
    \includegraphics[width=0.246\linewidth]{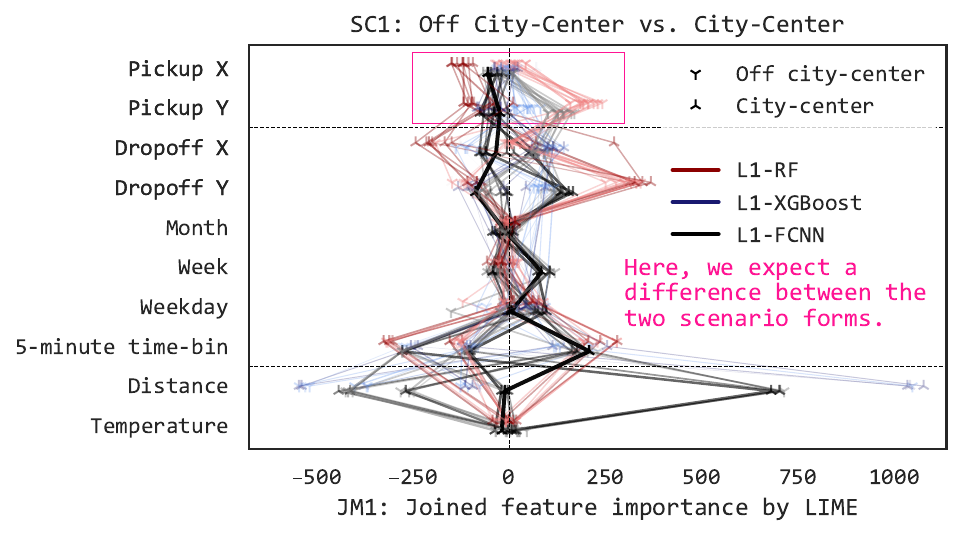}
    \hfill
    \includegraphics[width=0.246\linewidth]{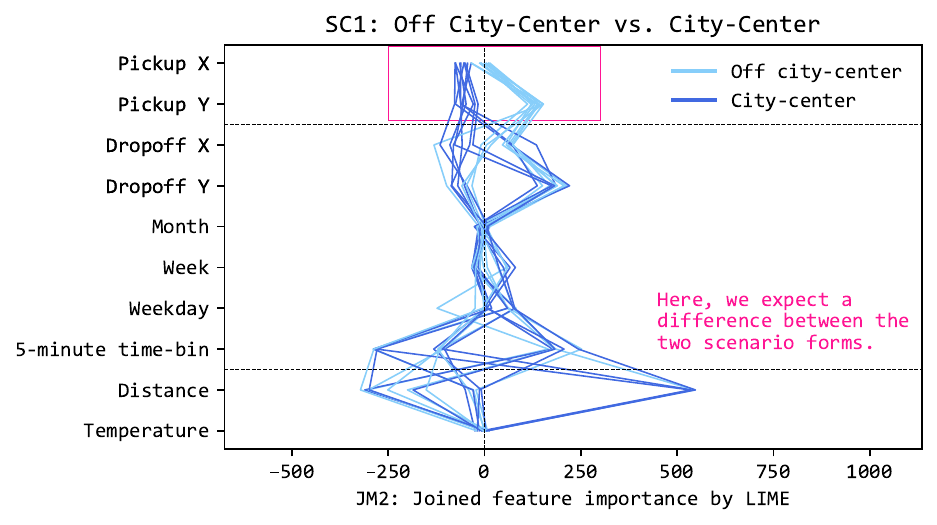}
    \hfill
    \includegraphics[width=0.246\linewidth]{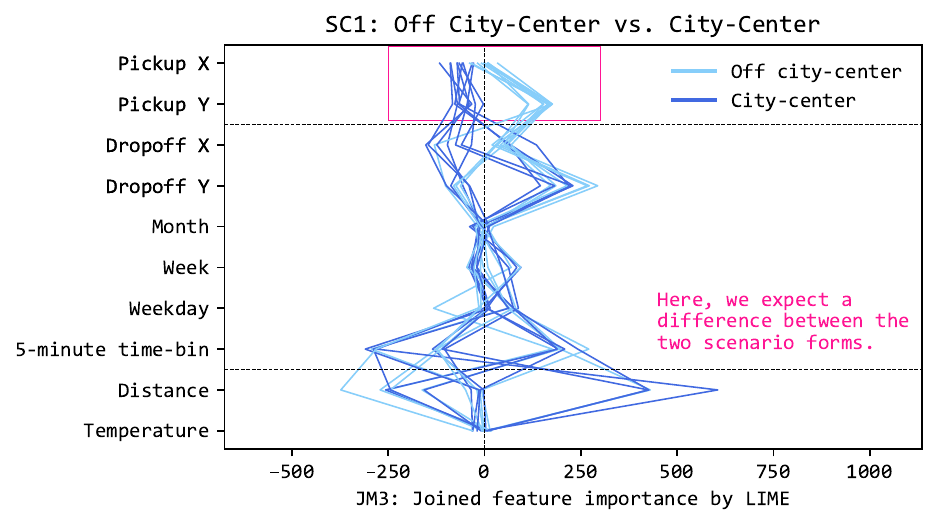}
\end{subfigure}
\begin{subfigure}[b]{\textwidth}
    \includegraphics[width=0.246\linewidth]{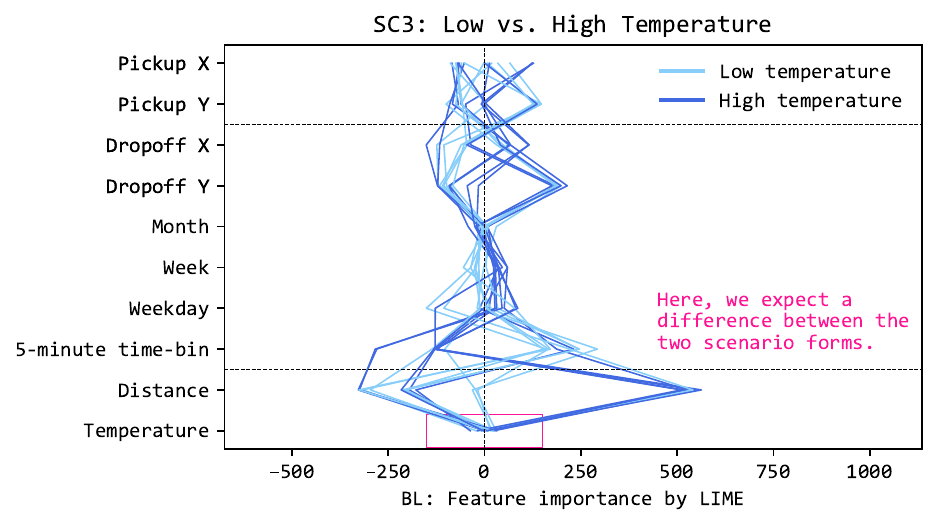}
    \hfill
    \includegraphics[width=0.246\linewidth]{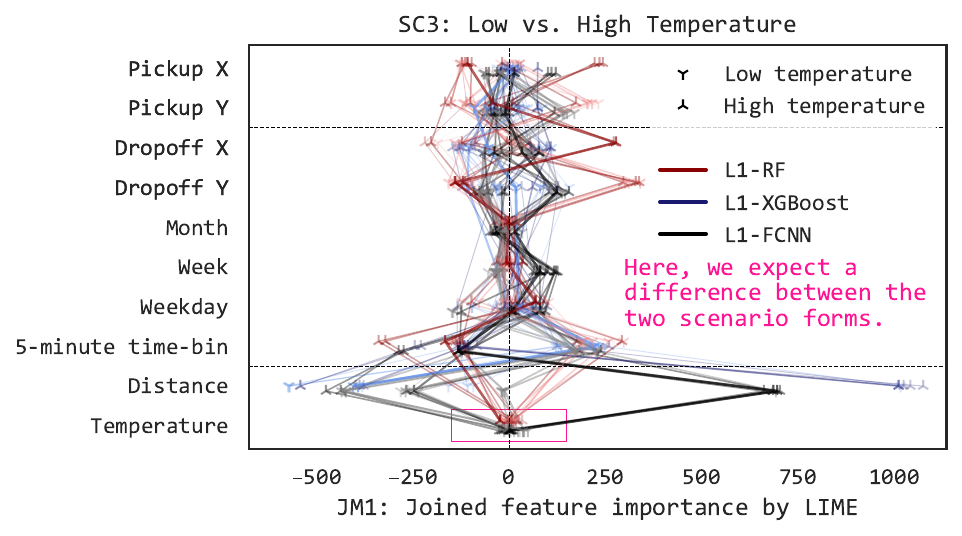}
    \hfill
    \includegraphics[width=0.246\linewidth]{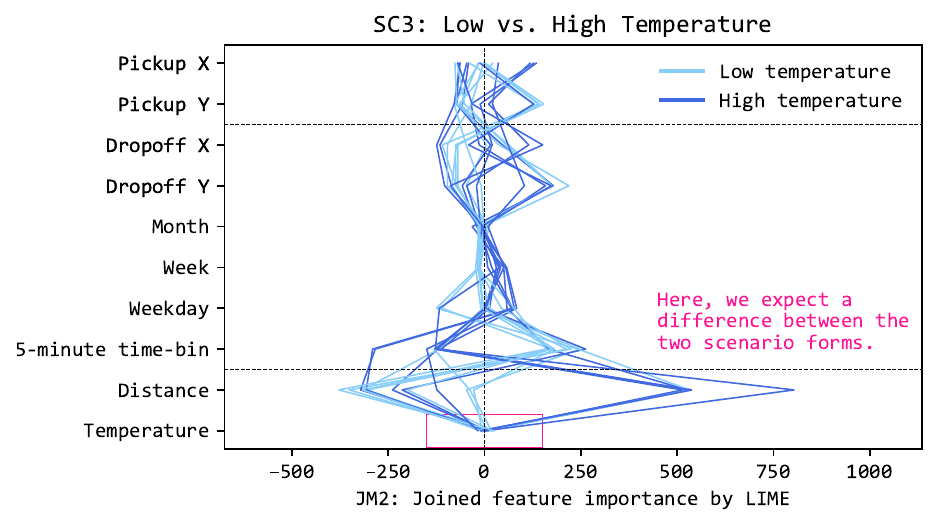}
    \hfill
    \includegraphics[width=0.246\linewidth]{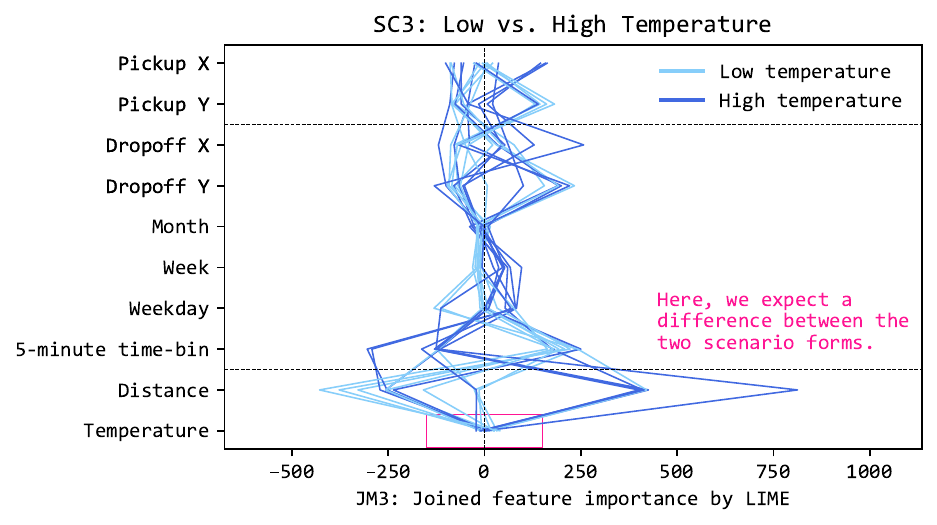}
\end{subfigure}
\begin{subfigure}[b]{\textwidth}
    \includegraphics[width=0.246\linewidth]{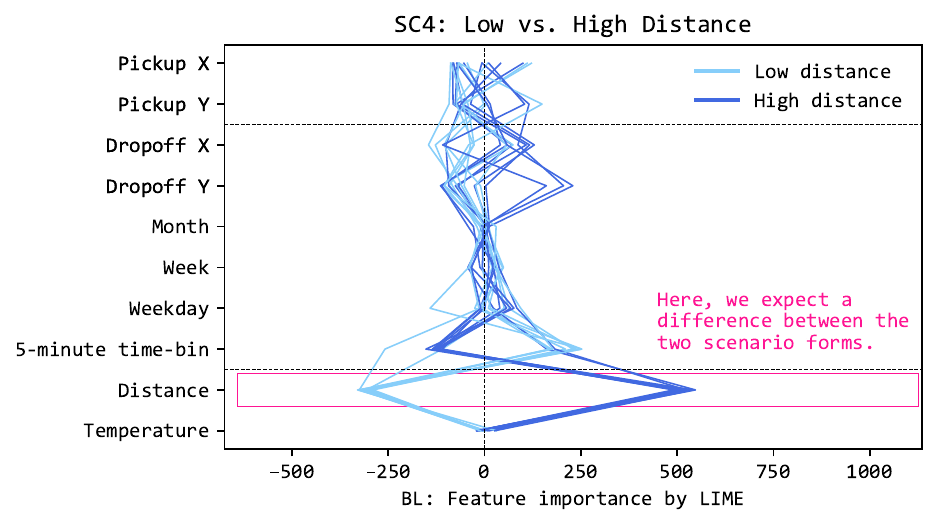}
    \hfill
    \includegraphics[width=0.246\linewidth]{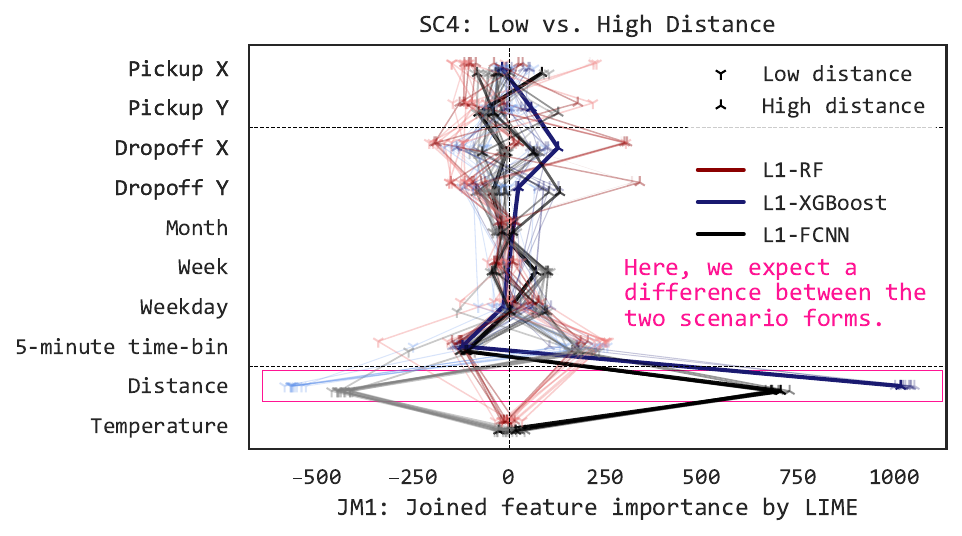}
    \hfill
    \includegraphics[width=0.246\linewidth]{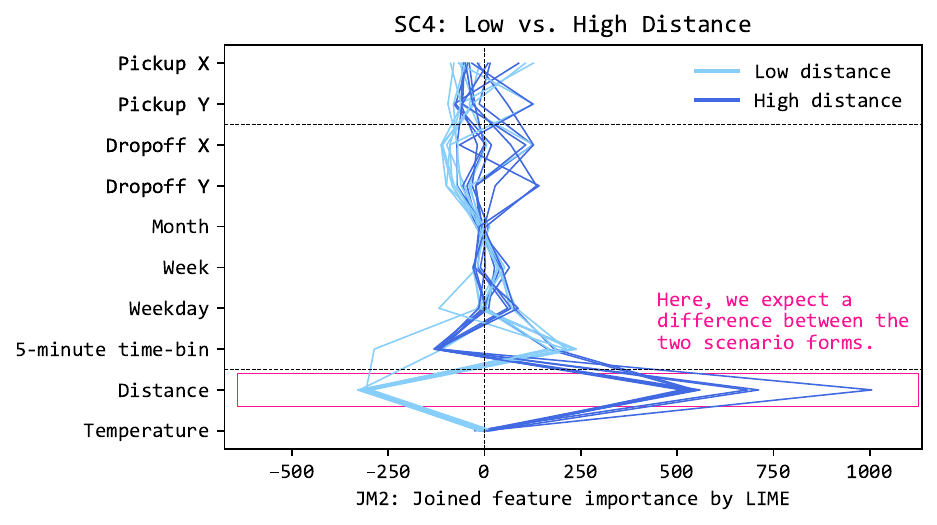}
    \hfill
    \includegraphics[width=0.246\linewidth]{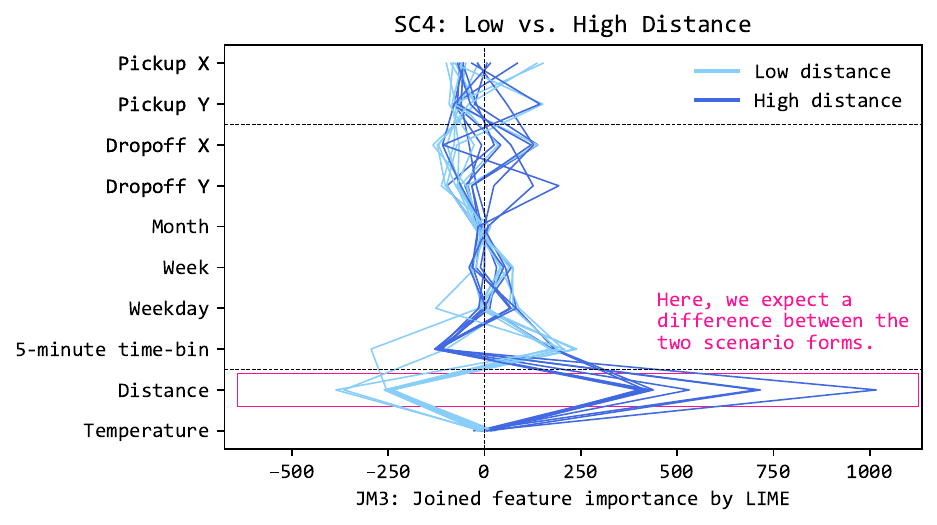}
\end{subfigure}
\caption{Content of Figure~\ref{fig:Lev2Scenario2LIMEv3} for SC1, SC3, and SC4.}
\label{fig:Lev2Scenario134LIMEv3}
\end{figure}

\begin{figure}[ht]
\centering
\begin{subfigure}[b]{\textwidth}
    \includegraphics[width=0.246\linewidth]{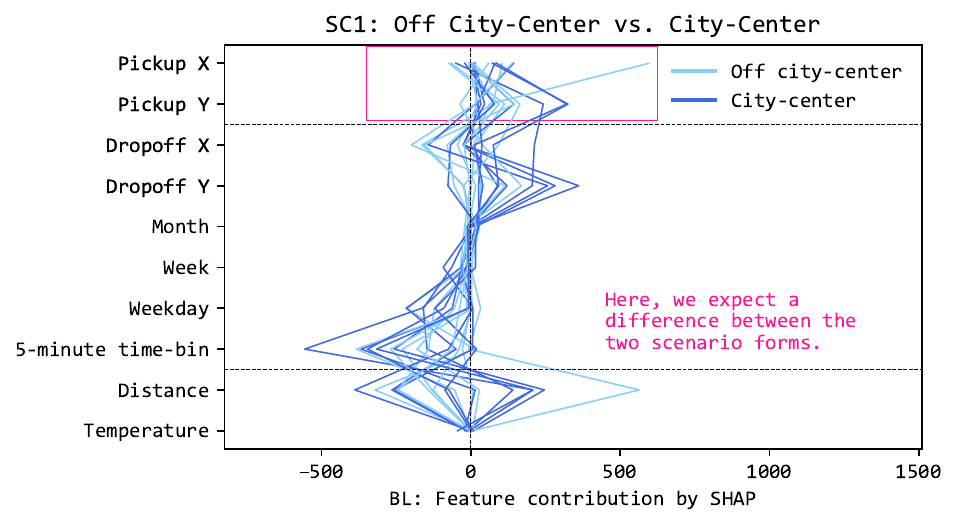}
    \hfill
    \includegraphics[width=0.246\linewidth]{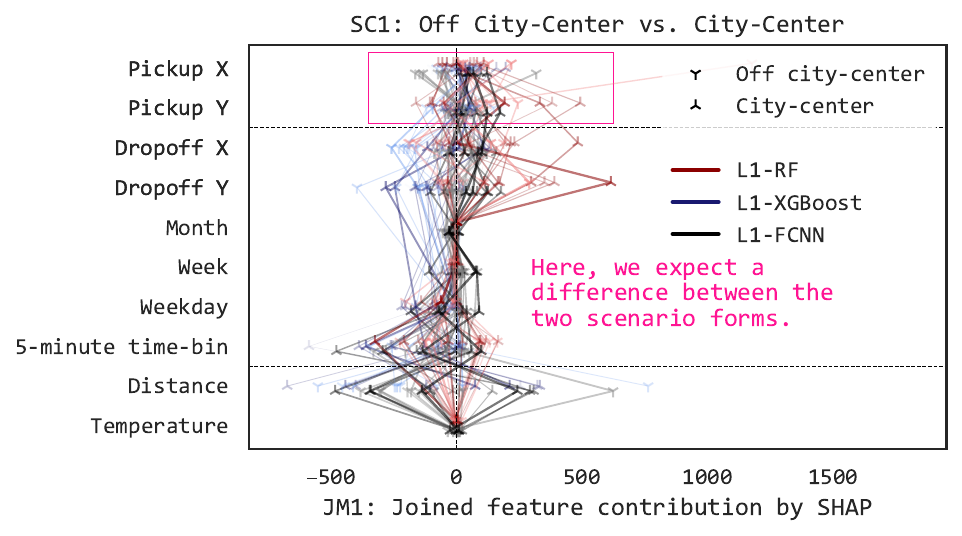}
    \hfill
    \includegraphics[width=0.246\linewidth]{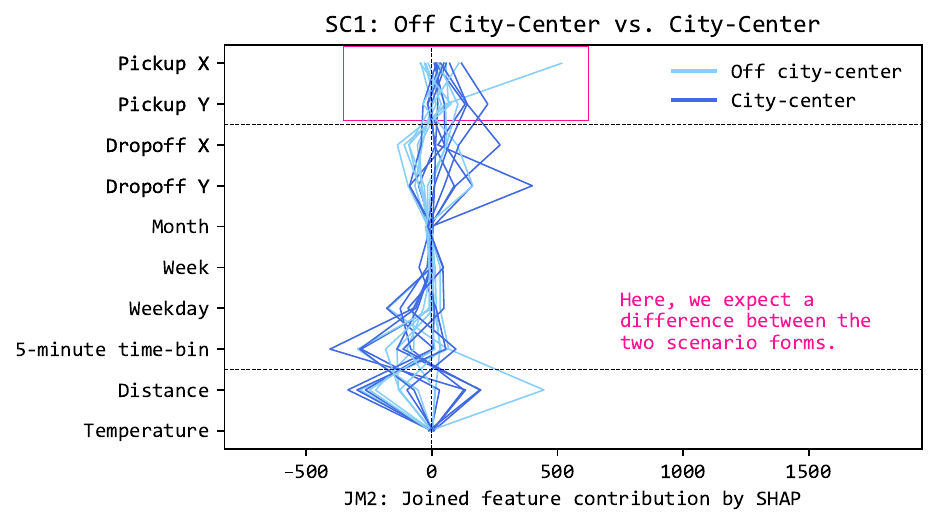}
    \hfill
    \includegraphics[width=0.246\linewidth]{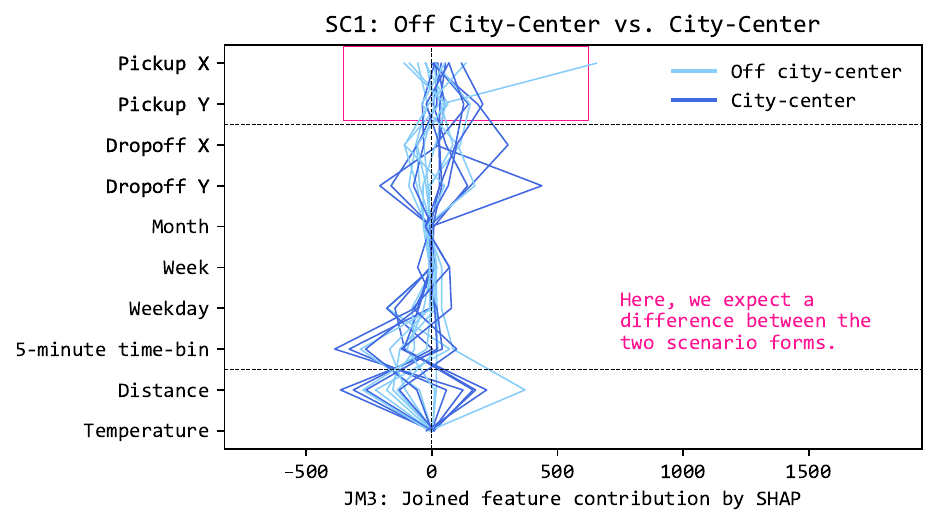}
\end{subfigure}
\begin{subfigure}[b]{\textwidth}
    \includegraphics[width=0.246\linewidth]{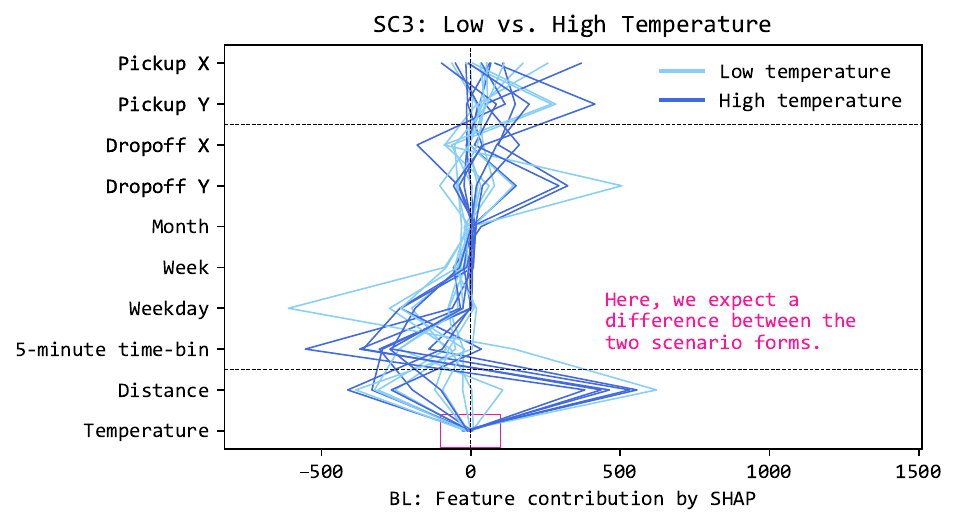}
    \hfill
    \includegraphics[width=0.246\linewidth]{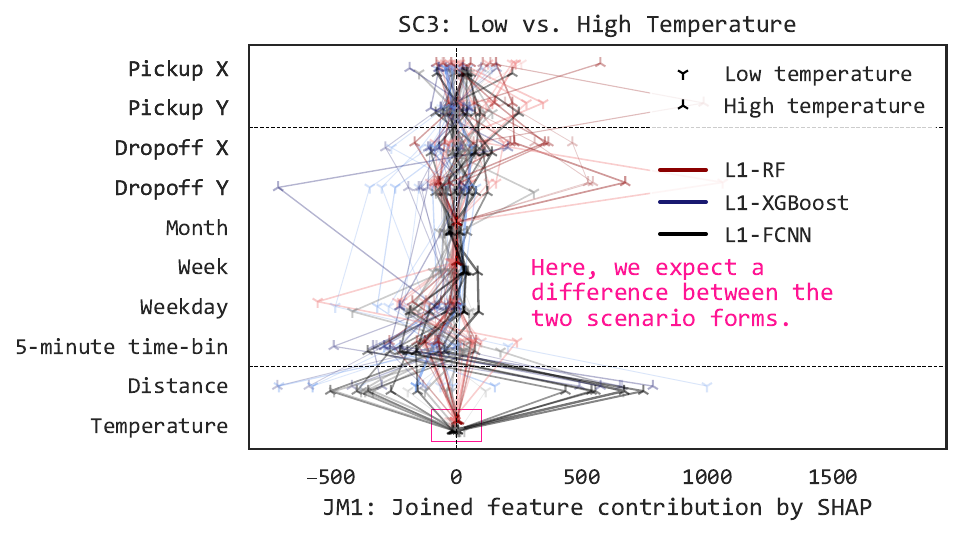}
    \hfill
    \includegraphics[width=0.246\linewidth]{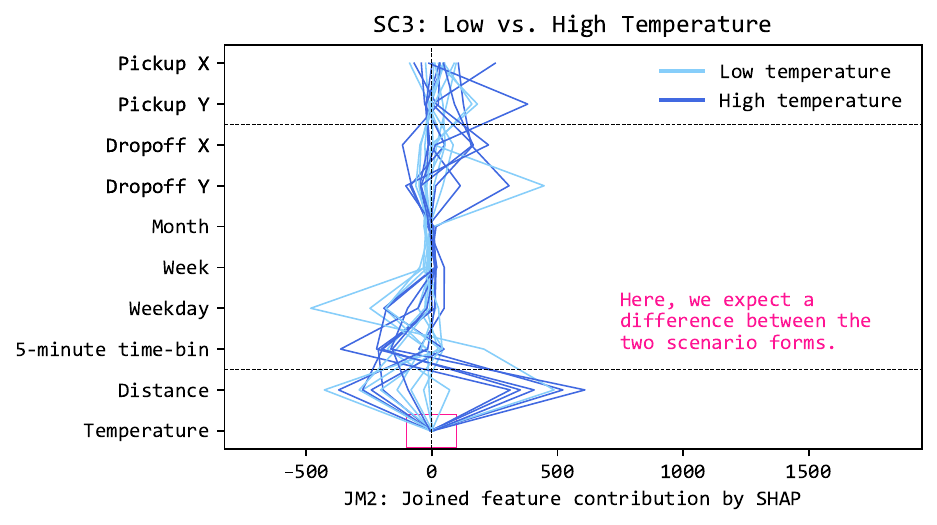}
    \hfill
    \includegraphics[width=0.246\linewidth]{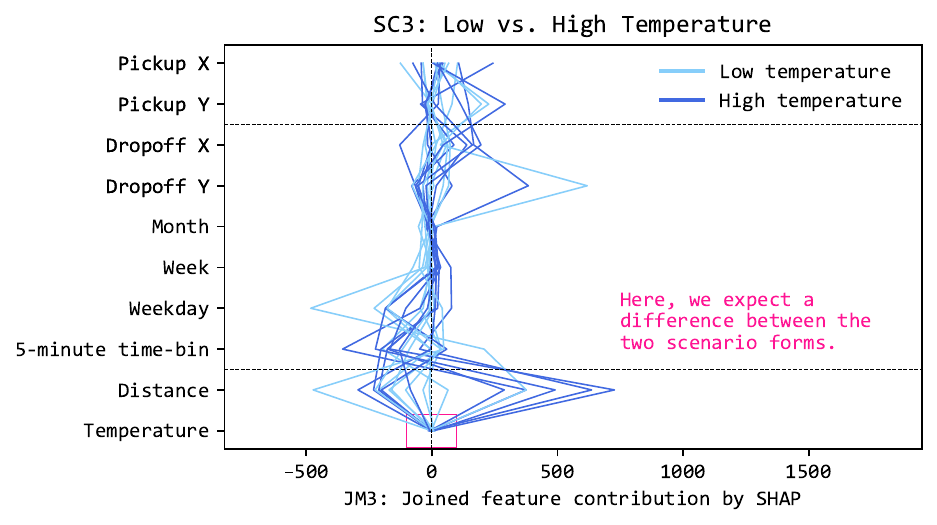}
\end{subfigure}
\begin{subfigure}[b]{\textwidth}
    \includegraphics[width=0.246\linewidth]{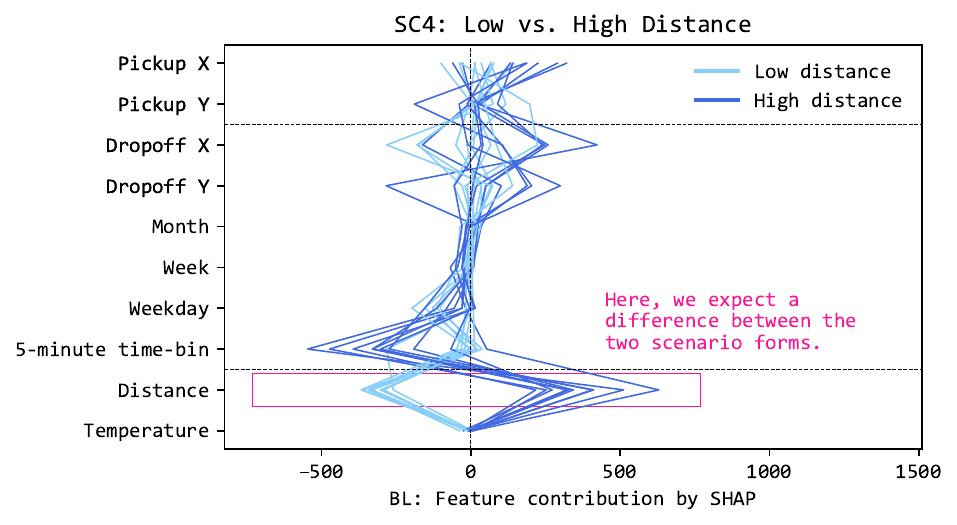}
    \hfill
    \includegraphics[width=0.246\linewidth]{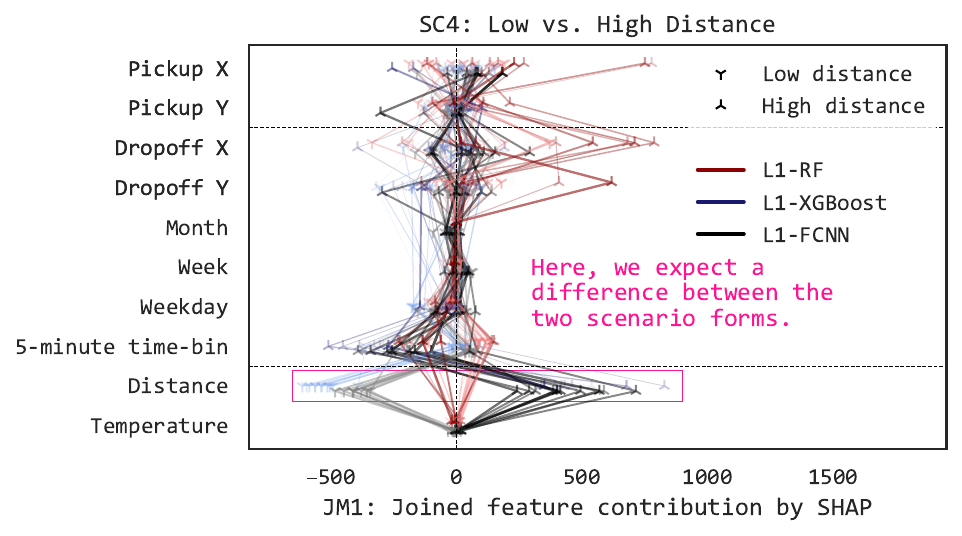}
    \hfill
    \includegraphics[width=0.246\linewidth]{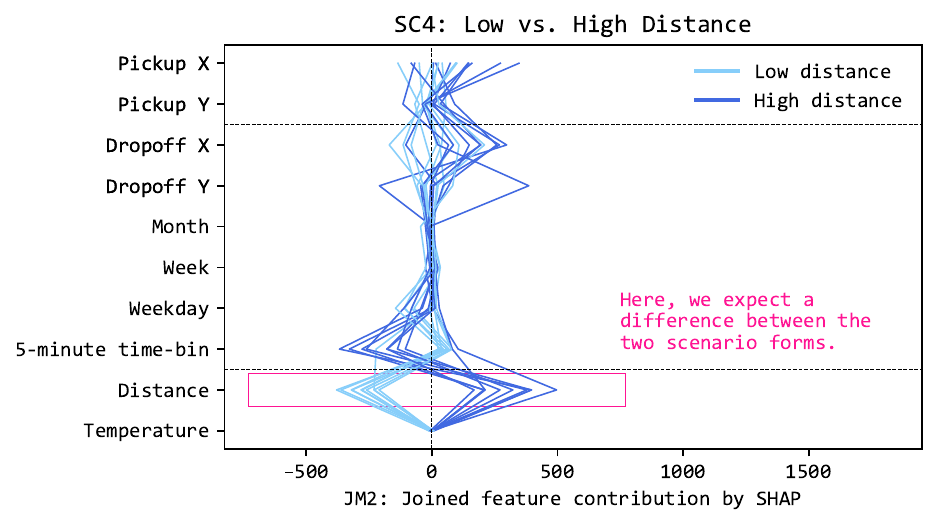}
    \hfill
    \includegraphics[width=0.246\linewidth]{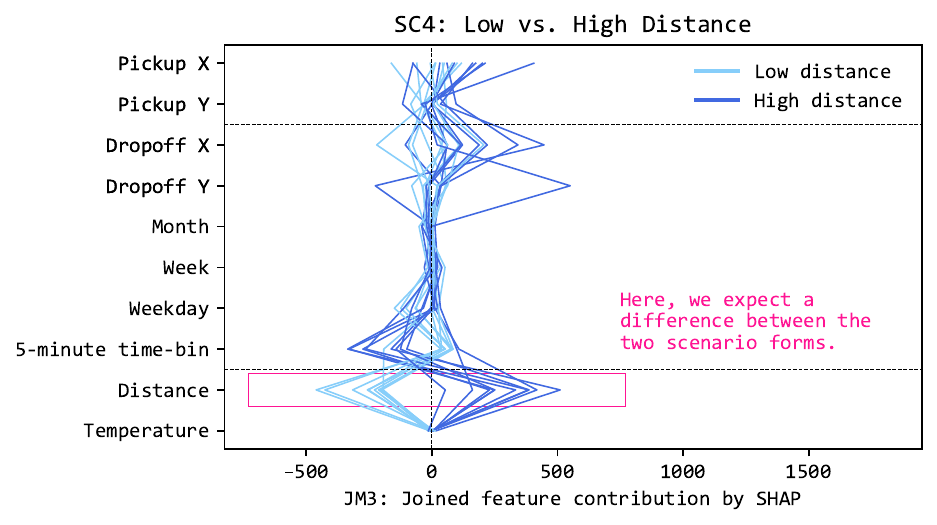}
\end{subfigure}
\caption{Content of Figure~\ref{fig:Lev2Scenario2SHAPv3} for SC1, SC3, and SC4.}
\label{fig:Lev2Scenario134SHAPEv3}
\end{figure}

\begin{figure*}[ht]
    \centering
    \includegraphics[width=\textwidth]{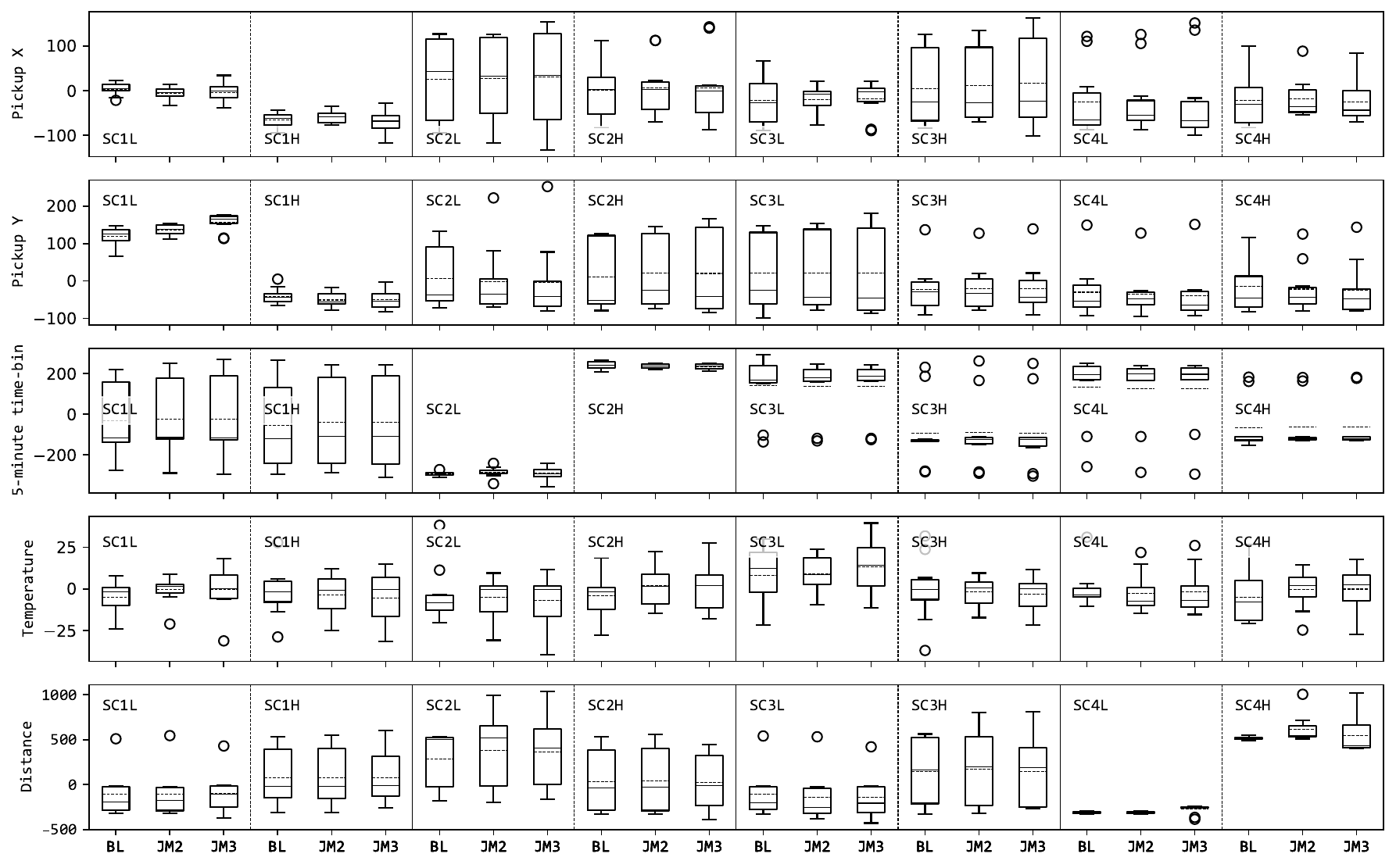}
    \caption{Content of Figure~\ref{fig:SHAPSCFeaturesBoxplots} for LIME}
    \label{fig:LIMESCFeaturesBoxplots}
\end{figure*}

In Figure~\ref{fig:Lev2Scenario134LIMEv3}, we visualize the \ac{LIME} explanations generated via the proposed joining methods for SC1, SC3, and SC4; in Figure~\ref{fig:Lev2Scenario134SHAPEv3}, we do the same for the \ac{SHAP} explanations. Figure~\ref{fig:LIMESCFeaturesBoxplots} shows the content of Figure~\ref{fig:SHAPSCFeaturesBoxplots} for LIME.

\clearpage
\printbibliography

\end{document}